%% file: sdm_arxiv.tex
\documentclass[twoside,leqno,twocolumn]{article}

\usepackage[letterpaper]{geometry}

\usepackage{ltexpprt}

\usepackage{hyperref}

\usepackage{amsmath,amsfonts}
\usepackage{algorithm}
\usepackage[noend]{algpseudocode}
\usepackage{array}
\usepackage{url}
\usepackage{graphicx}
\usepackage{cite}
\usepackage{booktabs}
\usepackage{cellspace}
\usepackage{eqparbox}
\usepackage{multirow}
\usepackage{subcaption}
\usepackage[shortlabels]{enumitem}
\input{math_commands.tex}

\long\def\comment#1{}

\begin{document}

\newcommand\relatedversion{}

\title{\Large Federated Koopman-Reservoir Learning \\ for Large-Scale Multivariate Time-Series Anomaly Detection\relatedversion}
\author{Long Tan Le\thanks{School of Computer Science, the University of Sydney (Email: \{long.le, suranga.seneviratne, nguyen.tran\}@sydney.edu.au)} 
\and Tung Anh Nguyen\thanks{School of Computer Science, the University of Sydney (Email: \{tung6100, hshu3770\}@uni.sydney.edy.au)}  
\and Han Shu\footnotemark[2]
\and Suranga Seneviratne\footnotemark[1] 
\and Choong Seon Hong\thanks{School of Computing, Kyung Hee University (Email:  cshong@khu.ac.kr)}
\and Nguyen H. Tran\footnotemark[1]}

\date{}

\maketitle


\fancyfoot[R]{\scriptsize{Copyright \textcopyright\ 2025 by SIAM\\
Unauthorized reproduction of this article is prohibited}}





\begin{abstract} \small\baselineskip=9pt
The proliferation of edge devices has dramatically increased the generation of multivariate time-series (MVTS) data, essential for applications from healthcare to smart cities. Such data streams, however, are vulnerable to anomalies that signal crucial problems like system failures or security incidents. Traditional MVTS anomaly detection methods, encompassing statistical and centralized machine learning approaches, struggle with the heterogeneity, variability, and privacy concerns of large-scale, distributed environments. In response, we introduce \textsc{FedKO}, a novel unsupervised Federated Learning framework that leverages the linear predictive capabilities of Koopman operator theory along with the dynamic adaptability of Reservoir Computing. This enables effective spatiotemporal processing and privacy-preserving for MVTS data. \textsc{FedKO} is formulated as a bi-level optimization problem, utilizing a specific federated algorithm to explore a shared Reservoir-Koopman model across diverse datasets. Such a model is then deployable on edge devices for efficient detection of anomalies in local MVTS streams. Experimental results across various datasets showcase \textsc{FedKO}'s superior performance against state-of-the-art methods in MVTS anomaly detection. Moreover, \textsc{FedKO} reduces up to 8$\times$ communication size and 2$\times$ memory usage, making it highly suitable for large-scale systems.

\end{abstract}

\input{SDMSections_Full/introduction}

\input{SDMSections_Full/related_work}
\input{SDMSections_Full/background}
\input{SDMSections_Full/problem_formulation}

\input{SDMSections_Full/algorithms}
\input{SDMSections_Full/experiment}

\input{SDMSections_Full/conclusion}

\def\url#1{}
\bibliographystyle{IEEEtranNoURL}
\bibliography{sdm_references} 

\appendix
\input{SDMSections_Full/appendix1}

\end{document}

%% file: math_commands.tex
\usepackage{amsmath,amsfonts,bm}

\def\eqref#1{(\ref{#1})}

\def\1{\bm{1}}
\DeclareUnicodeCharacter{FB01}{fi}



%% file: SDMSections_Full/introduction.tex
\section{Introduction}

The ubiquity of large-scale systems marks a transformative shift, enabling intelligent devices to communicate and make autonomous decisions with minimal human intervention. This advancement has spurred an increase in edge data generation, primarily in the form of multivariate time series (MVTS) that reflect dynamic, evolving processes in real-world settings. Such data, characterized by their spatial and temporal nature, are essential for large-scale applications. However, unusual patterns in these data streams, signaling anomalies, often indicate serious issues like system malfunctions and security breaches. Thus, timely detection and analysis of these anomalies are essential to mitigate potential damage and ensure system reliability and safety. 


Conventional MVTS anomaly detection (MTAD) typically employs statistical and machine learning (ML) approaches. Statistical methods, such as autoregressive models and outlier detection algorithms~\cite{arima}, rely on assumptions of data stationarity and distribution properties. However, these methods struggle with the complexity and variability inherent in real-world edge data. On the other hand, ML methods have emerged as powerful alternatives, leveraging models like Long Short-Term Memory (LSTM)~\cite{smap, lstmae},  Autoencoder (AE)~\cite{vae1, ransyncoder, ae_sdm}, Generative Adversarial Networks (GANs)~\cite{gan1, usad, daemon}, Graph Neural Networks (GNNs)~\cite{gdn, gnn_icdm, grelen}, and Transformers~\cite{tranad, anomaly_transformer} to discern patterns and anomalies. These methods excel in identifying patterns and anomalies by modeling nonlinear relationships and capturing complex dependencies within MVTS data. Despite their effectiveness, ML-based approaches often necessitate centralizing data on a central server, raising privacy concerns in edge contexts. Additionally, their high complexity and computational demands can be challenging for communication and resource-constrained environments. 
Therefore, it is essential to develop efficient, resource-friendly MTAD frameworks for large-scale systems that ensure computational feasibility and privacy preservation.

Recently,\textit{ Federated Learning }(FL) has gained prominence as a notable distributed learning paradigm, allowing multiple devices to jointly train a shared model while keeping the data localized~\cite{fedavg}. This approach effectively tackles privacy and communication concerns, making it particularly advantageous in beneficial in large-scale environments where data security and reducing data transmission are paramount. Such advancement promises a new horizon in analyzing MVTS, where the power of collaborative, decentralized learning can be harnessed effectively, ensuring both privacy preservation and operational efficiency. Nevertheless, implementing FL with deep models poses challenges due to their computational intensity on resource-constrained devices.


In this paper, we propose \textit{\underline{Fed}erated \underline{K}oopman-Reserv\underline{o}ir Learning} (\textsc{FedKO}), an unsupervised learning framework jointly addressing \textit{privacy}, \textit{communication}, and \textit{computational} challenges in MTAD. \textsc{FedKO} harnesses the strengths of FL in conjunction with Koopman operator theory (KOT)~\cite{koopman} and Reservoir Computing (RC)~\cite{reservoir}. Particularly, KOT offers a robust linear approach to study time-varying systems, utilizing an infinite-dimensional linear operator to simplify the modeling of nonlinear dynamics and complex behavior of MVTS. Meanwhile, RC excels at rapidly processing and analyzing intricate data patterns through its ability to generate time series representations. This synergy not only enhances the ability to analyze and detect complex spatiotemporal patterns in MVTS but also reduces the complexity of training and computation processes. Consequently, \textsc{FedKO} emerges as an effective tool for large-scale data analysis and streaming, ideal for distributed environments where limited computational resources and stringent data privacy are critical factors.
The main contributions are summarized as follows:
\begin{itemize}[leftmargin=15pt, topsep=0pt]
    \item We propose \textsc{FedKO}, an unsupervised, privacy-preserving and resource-friendly FL framework, utilizing RC and KOT for efficient MTAD. \textsc{FedKO} reinterprets nonlinearities in distributed MVTS data into linear analysis, streamlining MTAD and reducing communication and computational demands.
    \item We develop a novel model, (\textsc{ReKo}), as part of \textsc{FedKO} to enhance anomaly detection. This involves creating a bi-level optimization framework to train a unified \textsc{ReKo} model across multiple devices using an FL algorithm, ensuring data privacy and addressing data heterogeneity in large-scale systems.
    \item Through experiments on diverse MVTS datasets, we demonstrate that \textsc{FedKO} outperforms other baselines in MTAD, highlighting its robustness to heterogeneous data distributions. Additionally, it significantly improves communication and memory efficiency, making it ideal for large-scale applications.
\end{itemize}

%% file: SDMSections_Full/related_work.tex
\section{Preliminaries and Related Works}
\label{sec:background}

\subsection{Multivariate Time-Series Anomaly Detection:}
A diverse array of ML techniques has been investigated for MTAD. LSTM-based models are effective in capturing temporal dependencies in context-rich anomaly sequences~\cite{smap, lstmae} but are often computationally intensive. Autoencoder-based reconstruction methods~\cite{vae1, ransyncoder, ae_sdm} excel at detecting deviations by compressing and reconstructing normal data but struggle with subtle anomalies. GAN-based models~\cite{usad, daemon} enhance detection robustness through data synthesis but face challenges with training stability. Recently, GNNs have gained prominence in analyzing time-series data with relational structures~\cite{gdn, gnn_icdm, grelen}. These models facilitate the modeling of data point interdependencies but face scalability challenges due to computational intensity. Meanwhile, Transformer-based models~\cite{anomaly_transformer, tranad} employ self-attention mechanisms for detailed analysis of long-term patterns. They enable parallel processing but remain computationally costly for lengthy and high-dimensional sequences. Despite offering advanced capabilities, these methods face challenges like \textit{computational complexity} and \textit{data sensitivity}, underscoring the need for more resource-efficient approaches.

\subsection{Federated Learning:} 
FL enables training ML models across numerous devices while maintaining data privacy. In FL, models are are trained locally on each device and then aggregated on a central server to collaboratively improve performance~\cite{Konecny2015}. 
Since the FedAvg method's inception~\cite{fedavg}, various FL adaptations have been developed, advancing large-scale applications in many areas~\cite{fl_iot, fl_iot_sdm}, including MVTS analysis for supervised classification~\cite{flames2graph} or forecasting~\cite{fedmssa}. However, FL faces \textit{communication and computation challenges} due to exchanging large model parameters and training demands on resource-limited devices. Efficient strategies have been proposed to alleviate these burdens like quantization~\cite{pmlr-v108-reisizadeh20a} -- reducing precision of model weights to simplify computations, compression~\cite{com1} -- reducing the model size to make it more manageable to transmit, and adaptive communication~\cite{comadaptive} -- modifying data exchange dynamically based on different factors.  Nonetheless, these methods often introduce additional computational overhead, making them difficult to scale efficiently.

%% file: SDMSections_Full/background.tex

\begin{figure}[t]
 	\vspace{-15pt}
	\centering    
	\includegraphics[width=0.85\linewidth]{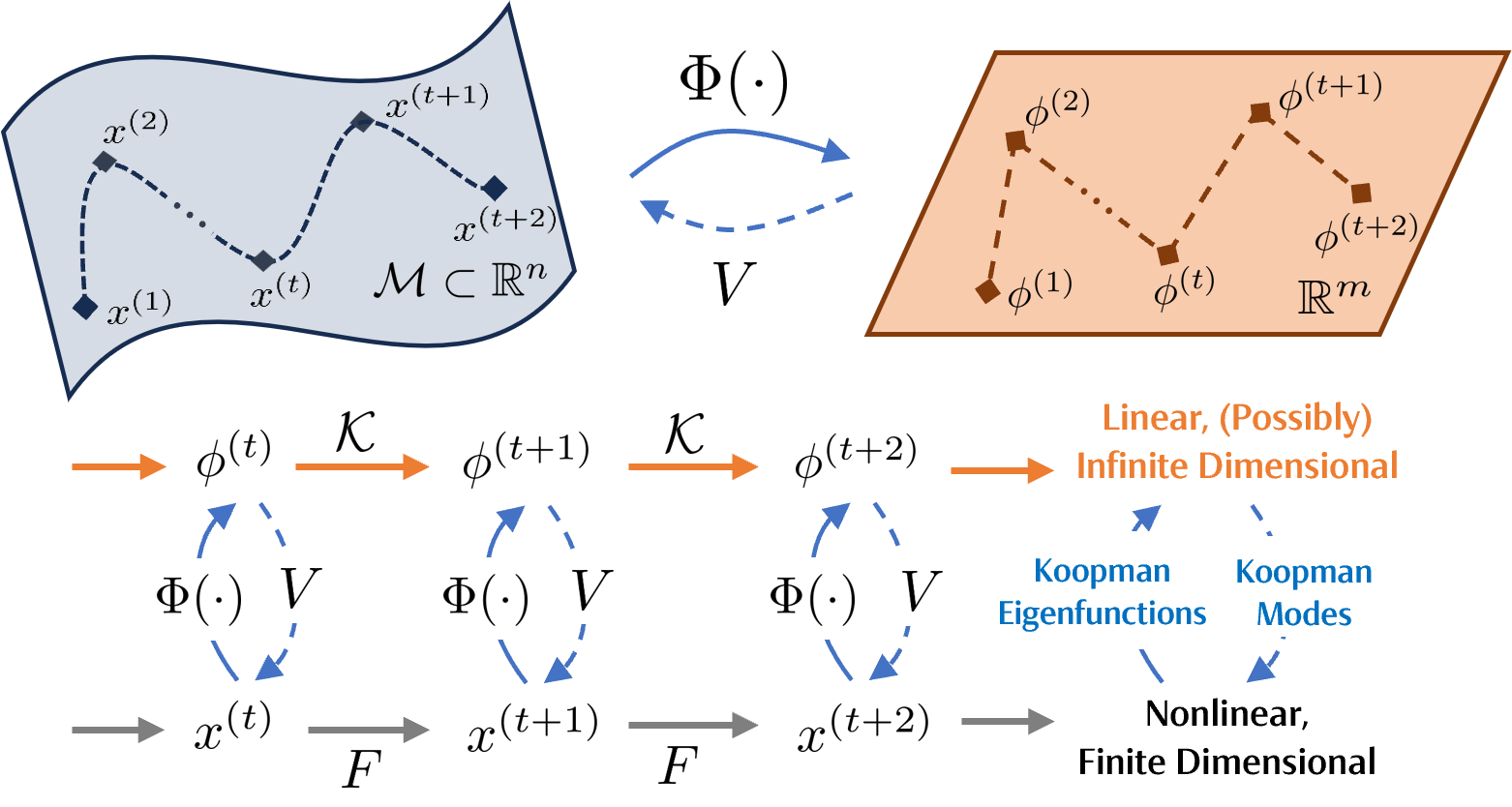}
 	\vspace{-5pt}
    \caption{Transformation of nonlinear dynamics to a higher-dimensional linear space via Koopman eigenfunctions \(\Phi(\cdot): \mathbb{R}^n \rightarrow \mathbb{R}^m\) (\(m > n\)), where each state \(x^{(t)} \in \mathbb{R}^n\) is mapped to an observable \(\phi^{(t)} \in \mathbb{R}^m\) that evolves linearly over time with Koopman operator \(\mathcal{K}\).} 
    \vspace{-15pt}
	\label{fig:koopman}
\end{figure}

\begin{figure*}[t]
 	\vspace{-15pt}
	\centering    
	\includegraphics[width=0.8\linewidth]{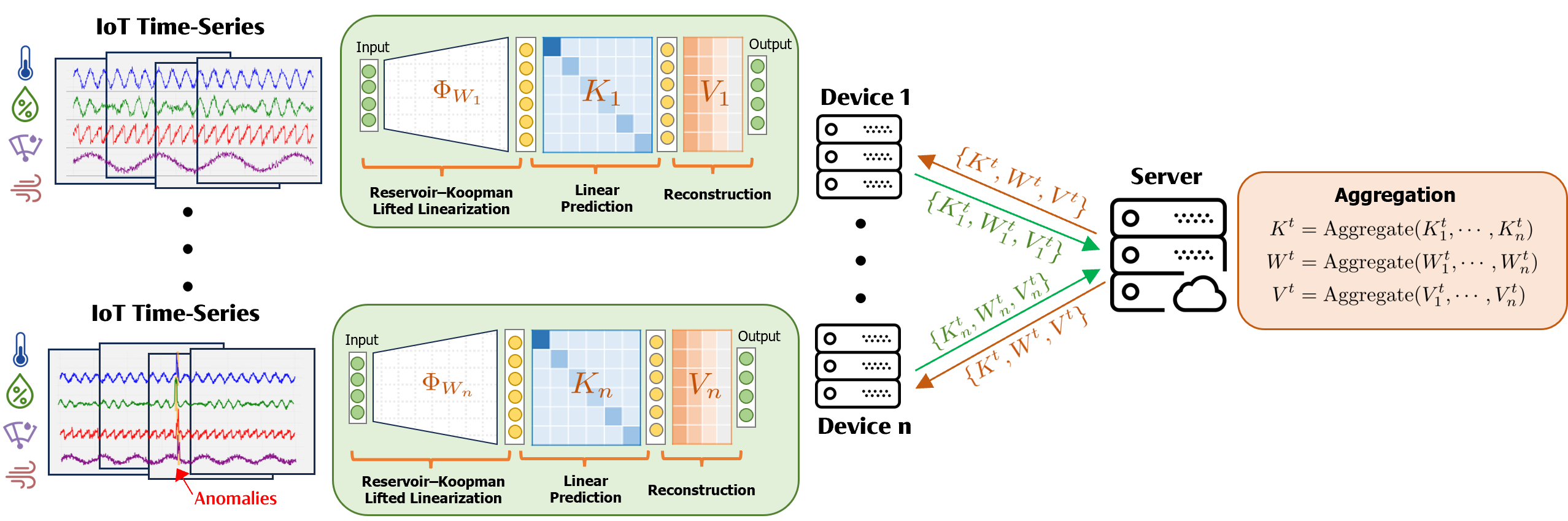}
	\caption{Illustration of \textsc{FedKo} where devices jointly learning a global \textsc{ReKo} model without data sharing.}
	\label{fig:gm}
 \vspace{-10pt}
\end{figure*}
\subsection{The Koopman Operator Theory:}
In nonlinear dynamical systems, such as fluid dynamics~\cite{fluid_dynamics}, the analysis typically revolves around state changes over time, described by nonlinear governing equations
$x^{(t+1)} = F(x^{(t)})$,
where $x \in \mathcal{M} \subset \mathbb{R}^n$ represents the states of system on a smooth $n$-dimensional manifold $\mathcal{M}$, and $F: \mathcal{M} \rightarrow \mathcal{M}$ is a nonlinear mapping that transitions the state $x^{(t)}$ to $x^{(t+1)}$ over a discrete interval. Koopman's theory~\cite{koopman} offers a powerful framework for \textit{stability analysis} of such systems by reinterpreting finite-dimensional nonlinear dynamics as a infinite-dimensional, yet linear, perspective. Central to this formalism is the Koopman operator $\mathcal{K}$, an infinite-dimensional linear operator. It acts on a specially defined observable space $g(x): \mathcal{M} \rightarrow \mathbb{R}^n$ of the state vector $x$, evolving over time according to a linear transformation as follows.
$$
    g(x^{(t+1)}) = \mathcal{K}g(x^{(t)}) = g \circ F(x^{(t)}).
$$
This linearization hinges on the system's observables, significantly simplifying the analysis of nonlinear dynamics. In practice, the function $g(x)$ often represents the state itself, i.e, $g(x) = x$~\cite{kotop}. Therefore, for subsequent analysis in this work, we will directly use $x$ to represent the system's state when discussing KOT. 

\vspace{5pt}
\noindent\textbf{Finite-Dimensional Approximation of $\mathcal{K}$:} Despite KOT's robust foundation, practical use is hindered by its infinite-dimensional nature. Recently, data-driven techniques~\cite{dmd1, dmd2, edmd1, edmd2} have paved the way for effective Koopman operator finite approximation. In addition, ML-based approaches to KOT~\cite{aekoopman, aekoopman2} leverage neural networks to approximate such operators, capturing non-linear dynamics more effectively. 




\vspace{5pt}
\noindent\textbf{Koopman Invariant Subspaces:}
By approximating the operator $\mathcal{K}$ as $K \in \mathbb{R}^{m \times m}$, where $m > n$ signifies a transition to a higher-dimensional space, one can leverage its linearity to facilitates the eigendecomposition: ${K}\Phi = \Phi\Lambda $. Here, $\Phi: \mathbb{R}^{n} \rightarrow \mathbb{R}^{m}$ denotes a set of Koopman eigenfunctions, and $\Lambda := \text{diag}(\lambda_1, \cdots,  \lambda_m)$ contains the corresponding eigenvalues.
This process not only linearizes the nonlinear dynamics but also establishes an intrinsic coordinate system through the eigenfunctions, offering a principled framework for understanding the system's behavior. It further identifies \textit{invariant subspaces} to illuminate the system’s dynamics linearly, \textit{enabling the reconstruction of the original state $x$ from this expanded space}, s.t., $x \approx \Phi(x) V$, with $V \in \mathbb{R}^{m \times n}$ representing the reconstruction matrix. This decomposition facilitates predictions as:
\begin{equation}
    x^{(t+1)} \approx \Phi(x^{(t+1)}) V =  K \Phi(x^{(t)}) V = \Phi(x^{(t)}) \Lambda V ,
    \label{eq:reconstruction}
\end{equation}
allowing the use of mode decomposition for forward-looking predictions, as depicted in Fig.~\ref{fig:koopman}. To construct Koopman eigenfunctions, various approaches have been proposed such as optimization-based~\cite{kotop1, kotop} and ML-based methods~\cite{kotop2}.

%% file: SDMSections_Full/problem_formulation.tex
\section{Federated Koopman-Reservoir Learning}
\label{sec:problem_formulation}

\subsection{Problem Description:}
Consider a large-scale system consisting of $N$ devices and a central server. Each device acquires multivariate temporal data, forming an MVTS dataset. This data contributes to a \textit{state} of the system, evolving over time through a complex, often nonlinear interplay of real-world factors. 
Under \textit{normal operation conditions}, these MVTS patterns are typically \textit{consistent} and \textit{predictable}. However, the system can be compromised by various vulnerabilities. Thus, analyzing these MVTS to promptly pinpoint anomalies that diverge from expected norms is crucial. 
Traditional centralized MTAD methods~\cite{lstmae, gdn, tranad, anomaly_transformer}
often pose privacy risks and demand substantial communication, computational and memory resources -- significant challenges for resource-constrained environments.
We, therefore, propose \textsc{FedKO} to jointly tackle these challenges. As shown in Fig.~\ref{fig:gm}, \textsc{FedKO} synergies FL with the linear dynamical system modeling capability and the computational efficiency of Koopman operator theory and Reservoir Computing. The core aim is to establish an efficient, privacy-preserving framework for detecting anomalies in MVTS data based on the fusion of RC with KOT, as detailed below.
\subsection{Reservoir-Koopman Model:}
\label{sec:model}
We first propose a novel parametric model, namely Reservoir-Koopman model (\textsc{ReKo}), for spatiotemporal processing MVTS. It comprises three key components: a \textit{Reservoir-Koopman lifted linearization} component $\Phi$; a \textit{learnable Koopman operator} $K$; and a \textit{learnable reconstruction matrix} $V$. These components function collaboratively and form the following predictive framework as follows:
\begin{subequations}
\begin{align}
    \phi^{(t)} & = \Phi(x^{(t)})\label{eq:1} \quad \text{ (Lifted Linearization)} \\
    \phi^{(t+1)} & = K \phi^{(t)} \qquad \text{(Linear Prediction)}  \label{eq:2}\\
    x^{(t+1)} & =V \phi^{(t+1)}  \quad \text{(Reconstruction)}  \label{eq:3}
\end{align}
\label{eq:main}
\end{subequations}
where $x^{(t)} \in \mathbb{R}^n$ is the MVTS sample at time step $t$ and $\phi^{(t)} \in \mathbb{R}^m$ ($m > n$) represents the corresponding transformed state by $\Phi(x^{(t)})$.

\subsubsection*{Koopman Lifted Linearization via Reservoir Computing:}
To tackle nonlinearities and spatiotemporal dependencies of MVTS, we initiate with the process of \textit{lifted linearization}, which involves transforming MVTS nonlinear dynamics into a linearizable framework through \textit{an approximation of Koopman eigenfunctions}. This approach redefines variable interrelations within MVTS's invariant subspace, facilitating their analysis via KOT.  
Towards this goal, we propose to employ \textit{Reservoir Computing} (RC)~\cite{esn, reservoir}, a cutting-edge approach for time-series embedding~\cite{reservoir2}, to create a Reservoir-Koopman lifting component $\Phi(\cdot): \mathbb{R}^n \rightarrow \mathbb{R}^m$. The gist of $\Phi$ is to \textit{lifts} the input data $x^{(t)} \in \mathbb{R}^n$ into a higher-dimensional state $\phi^{(t)} \in \mathbb{R}^m$ ($m > n$), where the dynamic is approximately linear, articulated in Eq.~\ref{eq:1}.





\begin{figure}[t]
\centering
\includegraphics[width=0.85\linewidth]{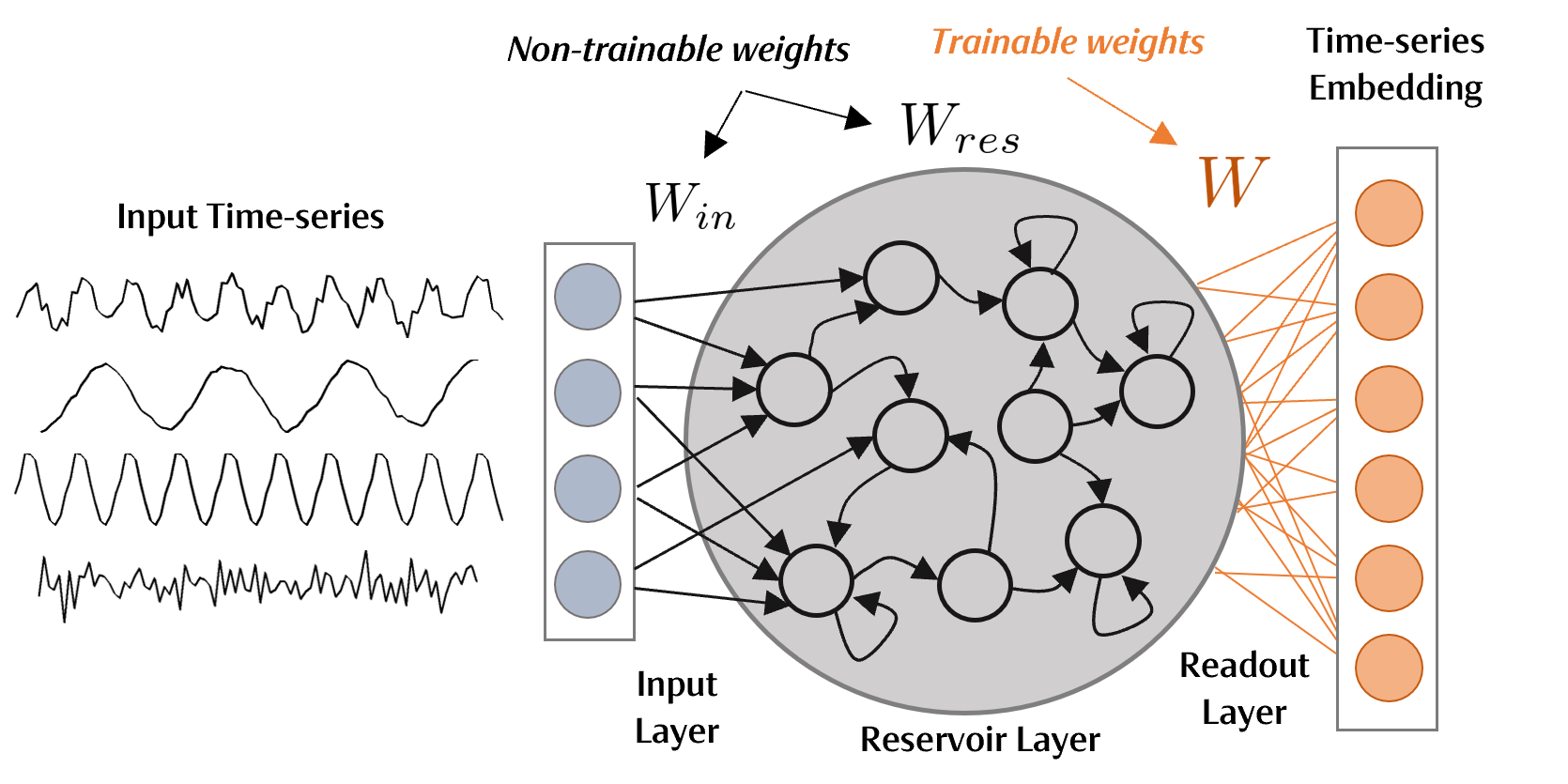}
\caption{The architecture of lifted linearization component $\Phi$ for \textsc{ReKo}, consisting of an \textit{untrainable reservoir} $(W_{in}, W_{res})$ for mapping inputs into a high-dimensional space and a \textit{trainable readout layer} $W$ for pattern analysis from the reservoir states.}
\label{fig:reservoir}
\vspace{-10pt}
\end{figure}
As depicted in Fig.~\ref{fig:reservoir}, the RC-based component $\Phi$ employs a three-layer architecture: an input layer with the weight $W_{in}$, a reservoir layer with the weight $W_{res}$; and a readout layer with the weight $W$. Notably, $W_{in}$ and $W_{res}$ are often \textit{untrainable} and \textit{high-dimensional}, enabling complex transformations without requiring training, while $W$ is adaptable and constitutes the model's \textit{trainable} component.  The operation of $\Phi$ unfolds in two following primary stages:

\comment{
\textit{The Input Layer:}
Initially, each state $\mathbf{x^{(t)}}$ is transformed as it enters the reservoir. This transformation is mathematically represented by:
\begin{equation}
u^{(t)} = W_{in} \cdot x^{(t)}
\end{equation}
where $u_{(t)}$ signifies the transformed input, and $W_{in}$ represents the fixed input weight matrix.}

\textit{1. Reservoir Transformation:} At each time step $t$, an input vector $x^{(t)} \in \mathbb{R}^n$, where $n$ represents the number of variables in the MVTS, undergoes a transformation into the reservoir state as follows.
\begin{equation}
r^{(t)} = f(W_{res}r^{(t-1)} + W_{in}x^{(t)}).
\end{equation}

Here, $W_{in} \in \mathbb{R}^{d \times n}$ and $W_{res} \in \mathbb{R}^{d \times d} $ are both fixed, dictate the initial transformation and recurrent dynamics within the reservoir, respectively. The reservoir state $r^{(t)} \in \mathbb{R}^d$ is updated based on the current input $x^{(t)} \in \mathbb{R}^n$ and its previous state $r^{(t-1)}$ using a nonlinear function $f: \mathbb{R}^{d} \rightarrow \mathbb{R}^{d}$, capturing temporal dependencies and encapsulating the memory property of the system. 
Examples of nonlinear functions $f$ used in advanced RC architectures include those in Leaky Echo State Networks (ESN)~\cite{esn} or Reservoir Transformers~\cite{rctran}.

The fixed, random connectivity of the reservoir architecture enriches its internal state representation, thereby enhancing its ability to process spatiotemporal data~\cite{esn}. This architecture enables spatial multiplexing, where each neuron uniquely contributes to processing multivariate spatial information. The integration of spatial and temporal dimensions provides a multitude of \textit{degrees of freedom}, allowing the system to capture intricate patterns in MVTS effectively.
To choose parameters for these untrained layers, one can employ methods such as Bayesian optimization~\cite{optimalRC}, information theory~\cite{optimalRCE}, gradient and evolutionary optimization~\cite{optimalRCG}. 


\textit{2. Output Mapping:} The output $y^{(t)} \in \mathbb{R}^m$ (with $m > n$ is the approximated number of Koopman eigenfunctions), is derived from the reservoir's state: $y^{(t)} = h(W r^{(t)})$, 
where $h: \mathbb{R}^{m} \rightarrow \mathbb{R}^{m}$ the activation function for the readout layer. Unlike the reservoir, the readout's weight $W \in \mathbb{R}^{m \times d}$ is trainable, emphasizing that \textit{training RC-based models primarily involves minimize the error between the predicted outputs and the actual targets to obtain optimal $W$}. 

Within \textsc{FedKO}, the RC-based lifted linearization module $\Phi$ serves as a set of \textit{spatiotemporal filters}, simultaneously applied to the variables of MVTS to transform nonlinear input features into a higher-dimensional space. In this expanded space, complex, nonlinear relationships within the input data can often be represented as simpler, more linearly separable structures. This transformation is particularly appealing for the approximation of Koopman eigenfunctions, merging linear analytical techniques with the capability to unravel nonlinear dynamics in complex MVTS analysis.

\vspace{5pt}
\noindent\textbf{Linear Prediction via Koopman Operator:}
By applying $\Phi$ to approximate Koopman eigenfunction for MVTS data, the original input is transformed into a high-dimensional space, thereby achieving a structured linear representation crucial for adherence to KOT principles. 
This allows the Koopman operator $K$ to linearly forecast future states $\phi^{(t+1)}$ from the reservoir transformed state $\phi^{(t)}$ within this higher dimension space, as formulated by Eq.~\ref{eq:2}. Moreover, one can leverage the eigendecomposition of $K$ to predict a future data at time step $t$ from an initial state $\Phi(x^{(0)})$:
\begin{equation}
\Phi(x^{(t)}) \approx \underbrace{K(K(...(K\Phi(x^{(0)}))))}_{\text{$t$
nested Koopman operator}} =  \Phi(x^{(0)}) \Lambda^t.
\label{eq:kphi}
\end{equation}
Here, $\Lambda$, a diagonal matrix of eigenvalues of $K$, dictates the rate at which features evolve over time, highlighting the system’s dynamics through exponential scaling in the equation. This  not only allows for \textit{long-term predictions} within the linear space, but also balance between maintaining the data originality and achieving computational efficiency and predictive accuracy.

\vspace{5pt}
\noindent\textbf{Original Space Reconstruction via Koopman Mode Decomposition:}
In \textsc{ReKo}, predictions often extend into a higher-dimensional, yet linear space to capture complex MVTS patterns. The challenge then becomes reconstructing these high-dimensional predictions back into the original data space. Utilizing Koopman Mode Decomposition (KMD) in Eq.~\ref{eq:reconstruction}, the future value $x^{(t+1)} \in \mathbb{R}^{n} $ can be predicted based on the preceding value $x^{(t)} \in \mathbb{R}^{n}$, as indicated in Eq.~\ref{eq:3}. Furthermore, one can make long-term prediction for data $x^{(t)}$ from an initial $x^{(0)}$ in the original space as follows.
 \begin{equation}
x^{(t)}  \approx \Phi(x^{(0)}) \Lambda^t  V .
\label{eq:kmd}
\end{equation}
The reconstruction based on KMD ensures that the Koopman eigenfunctions $\Phi$ and Koopman modes $V$, which encapsulate the system's dynamics, remain invariant under the action of the Koopman operator. This is particularly valuable for forecasting and understanding the long-term behaviors of nonlinear systems.

\vspace{5pt}
\noindent\textbf{MVTS Anomaly Detection with \textsc{REKO}:} 
\textsc{ReKo} functions as a predictive model, adept at generating both \textit{one-step-ahead} and \textit{multi-step-ahead} forecasts from historical MVTS data using Eq.~\ref{eq:kmd}. Here, one can calculate anomalousness scores to measure the discrepancy between predicted \(\hat{x}^{(t)}\) by \textsc{ReKo} and actual \(x^{(t)}\) data, leveraging error metrics like Mean Squared Error (MSE).
A threshold \(\tau\) is then applied to the anomalousness score to classify data points as normal or anomalies, with scores above \(\tau\) indicating potential anomalies. 


The \textsc{ReKo} model enhances \textsc{FedKO} by leveraging the strengths of both RC and KOT, offering key benefits: (1) e\textit{nhanced MVTS spatial-temporal modeling} through high-dimensional data projection and linear refinement, (2) \textit{reduced complexity} via an untrained reservoir and efficient linear predictions, and (3) \textit{improved communication efficiency} by limiting updates during training. These features make \textsc{ReKo} ideal for edge computing in IoT, streamlining anomaly detection while conserving bandwidth and minimizing latency.

\begin{figure}[t]
	\centering    
	\includegraphics[width=0.9\linewidth]{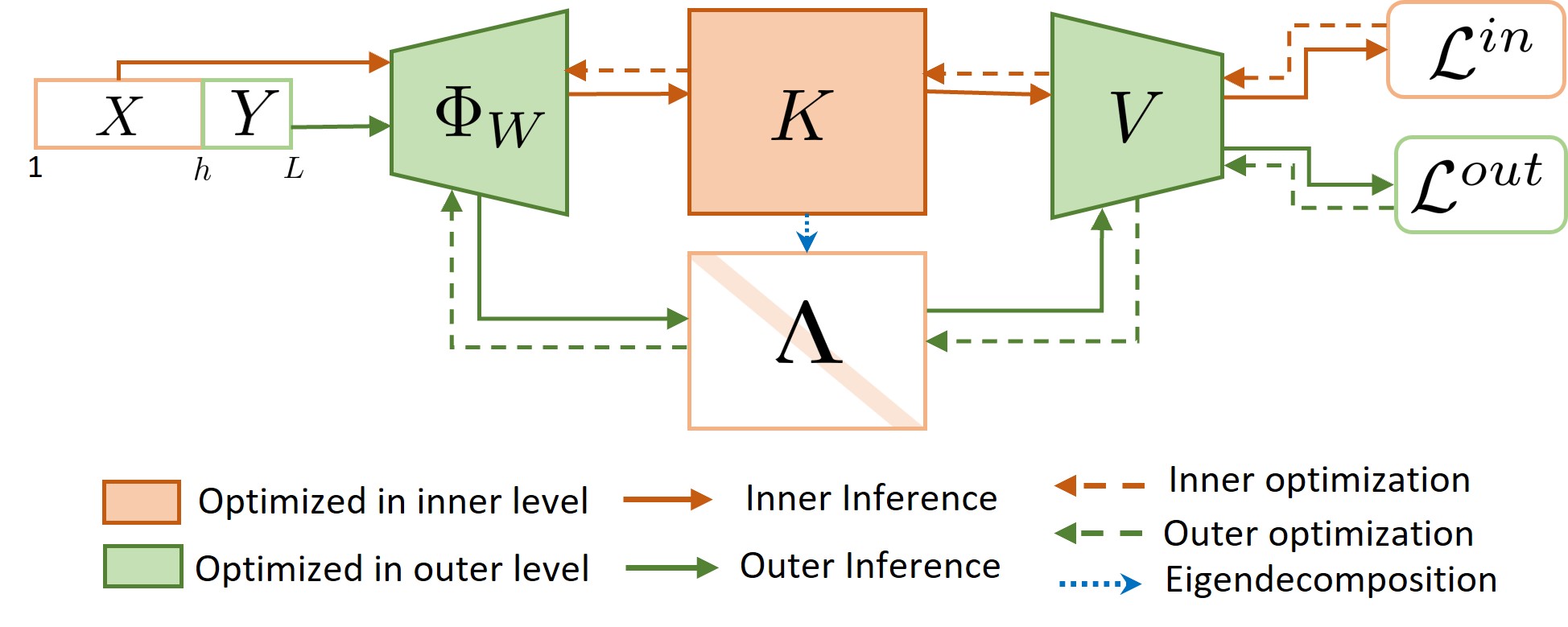}
	\caption{Overview of the \textsc{FedKO} bi-level framework.}
	\label{fig:bilevel}
    \vspace{-10pt}
\end{figure}

\subsection{\textsc{FedKO} -- Bi-level Optimization:}
\label{sec:optimization}

To learn a global \textsc{ReKo} model across multiple devices, we propose a novel federated optimization approach, aiming to leverage the local datasets at each device for facilitating efficient model learning while preserving data privacy. In particular, we consider a scenario where each device $i$ in the large-scale system possesses an MVTS dataset $\mathcal{X}_i \in \mathbb{R}^{n \times L}$, where $n$ and $L$ denote the number of variables and the length of the time series, respectively. These datasets are presumed to be collected under \textit{normal operational conditions}, characterized by consistent and predictable patterns over time. The goal for each device is to learn a local \textsc{ReKo} model including: (1) the Reservoir-Koopman eigenfunctions $\Phi_{W_i}: \mathbb{R}^{n} \rightarrow \mathbb{R}^m$ with its readout layer parameterized by $W_i$, (2) the Koopman operator $K_i \in \mathbb{R}^{m \times m}$, and (3) the reconstruction matrix $V_i \in \mathbb{R}^{m \times n}$. For this purpose, the local dataset $\mathcal{X}_i$ at each device is strategically divided into two segments: $X_i \in \mathbb{R}^{n \times s} \subset \mathcal{X}_i$, covering the interval  $[1, s]$ (with $s <L$); and $Y_i \in \mathbb{R}^{n \times (L- s + 1)} \subset \mathcal{X}_i$,  extending over $[s, L]$, as illustrated in Fig.~\ref{fig:bilevel}. This segmentation is instrumental in constructing a bilevel optimization framework to solve problem~\ref{eq:main} as follows.
\\
\textbf{\textit{Inner level}}:
\begin{align}
K_i^*\in \underset{K_i}{\text{argmin }} \Big\{ \mathcal{L}_i^{in}  & =  \|{\Phi}_{W_i}(X_i^{(t+1)}) - K_i {\Phi}_{W_i}(X_i^{(t)})\|_F^2 \nonumber \\ &  \qquad  + \|X_i^{(t+1)}-K_i {\Phi}_{W_i}(X_i^{(t)}) {V_i} \|_F^2 \Big\} \label{eq:inner}\\
\text{s.t. } \quad \rho(K_i) & < 1 \nonumber  
\end{align}
\textbf{\textit{Outer Level}}:
\begin{align}
 (W_i^*, V_i^*) \in \underset{W_i, V_i}{\text{argmin }} \Big\{ \mathcal{L}_i^{out} & = \sum_{j=0}^{L-s} \|y_i^{(j)} - \Phi_{W_i}(y_i^{(0)}) \Lambda_i^j  V_i\|_F^2 \Big\} \nonumber \\
\text{s.t. } \quad K_i^* \Phi_{W_i}(\cdot) & = \Phi_{W_i}(\cdot)\Lambda_i.  \label{eq:outer} 
\end{align}
Here, $\|\cdot\|_F$ and $\rho(K_i)$ denotes the Frobenius norm and the spectral radius of $K_i$, respectively. In the inner problem, the sequences $X_i^{(t)}$ and $X_i^{(t+1)}$, subsets of $X_i$, correspond to MVTS data over intervals $[t, t + b]$ and $[t+1, t + 1 + b]$ (one-time-step lag), with $b$ is a certain batch size.
In the outer problem, $y_i^{(j)} \in Y_i$, $j = 0, \cdots, L-s$,  represents a single time-series vector at time step $j$, and $\Lambda_i$ is the diagonal matrix of eigenvalues extracted from the eigendecomposition of $K_i^*$.  Based on the bilevel formulation, we introduce a novel FL algorithm for \textsc{FedKO}, to enable devices to collaboratively train a shared \textsc{ReKo} model without sharing local data. The key steps are outlined in Alg.~\ref{alg:fedko}, and are elaborated in detail as below. 

\subsubsection{Inner Problem:} 

\label{sec:inner}
The goal of Eq.~\ref{eq:inner} is to ensure accurate state transitions with the Koopman operator $K_i$ on linear space, adhering to the linearization principles of KOT. As depicted in Fig.~\ref{fig:bilevel}, it seeks to optimize $K_i$ in each device $i$ using part of its local dataset by minimizing the loss $\mathcal{L}_i^{in}$, consisting of two objective terms: (1) the linear prediction error between two consecutive lifted states transformed by $\Phi_{W_i}$; and (2) the linear prediction error between the actual next state and its predicted values reconstructed by $K_i$, $\Phi_{W_i}$, and $V_i$.

As in Alg.~\ref{alg:fedko}, at each training round $t$ of \textsc{FedKo}, a subset of devices will be selected and first solve the inner problem locally using gradient-based methods (e.g., ADAM) to obtain an optimal Koopman operator $K_i^{(t)}$ (lines 5-6). After that, $K_i^{(t)}$ are sent to the server for the first stage of aggregation, using Eq.~\ref{eq:update_K}, amalgamating the insights from all selected nodes, forming a preliminary global perspective of \textsc{ReKo} model (line 7).
\begin{equation}
\begin{split}
     K^{(t)} =  \beta K^{(t-1)} &+ (1-\beta)\frac{1}{|\mathcal{S}^{(t)}|} \sum\nolimits_{i \in \mathcal{S}^{(t)}} K_i^{(t)} \\
     \text{s.t.} \quad \rho(K^{(t)}) &< 1 
\end{split}
\label{eq:update_K}
\end{equation}
\textbf{Mitigating Non-IID via Weighted Aggregation:} Conventional averaging methods often struggle with slow convergence and suboptimal performance due to \textit{data heterogeneity}~\cite{fedavg}. In this work, we employ a \textit{weighted aggregation scheme} with a momentum factor $\beta$, incorporating prior model parameters into the aggregation process.
By blending a portion of the previous global weights with current local updates, it not only smooths out fluctuations inherent in non-IID datasets but also stabilizes and enhances convergence. 
Moreover, it addresses the client-drift issue by minimizing the divergence in local updates, thereby directing global model updates towards their optimal trajectory.

\vspace{5pt}
\noindent\textbf{Stabilizing Dynamics via Spectral Regularization}: To ensure the stability of \textsc{ReKo} model when training with normal data, we enforce\textit{ a spectral radius constraint $\rho(K_i) < 1$ on the Koopman operator $K_i$}. From KOT perspectives, $K_i$ with spectral radius below 1 denotes system stability, while values above 1 can signal potential instability~\cite{koopman}. By controlling $\rho(K_i)$, we enhance the model's ability to distinguish between true anomalies and regular patterns in MVTS, essential for reliable long-term behavior prediction. 
\begin{algorithm}[t]
    \small
	\caption{Federated Koopman-Reservoir Learning}
	\label{alg:fedko}
	\begin{algorithmic}[1]
        \State Initialize ($W_{in}$, $W_{res}$) for all training devices 
		\State Initialize $W_i^{(0)}$, $K_i^{(0)}$ and $V_i^{(0)}, \quad \forall i = 1, \dots, N$
		\For{$t = 1, \ldots, T$}
        \State Sample subset devices $\mathcal{S}^{(t)}$ 
		  \For {each device $i \in \mathcal{S}^{(t)}$ in parallel}
		\State Solve local inner problem \ref{eq:inner} to find $K_i^{(t)}$ \Statex \quad \qquad using gradient-based methods.
		\EndFor
        \State Server updates  $K^{(t)}$ using Eq.~\ref{eq:update_K} 
        \State Server broadcasts $K^{(t)}$ to all devices.
        
		\For {each node $i \in \mathcal{S}^{(t)}$ in parallel} 
        \State Extract $\Lambda^{(t)}_i$ from $K^{(t)}$ by  eigendecomposition
  		\State Solve local outer problem \ref{eq:outer} to refine $W_i^{(t)}$ and
        \Statex \quad \qquad $V_i^{(t)}$ using gradient-based methods.
		\EndFor
        \State Server updates $W^{(t)}$  using Eq.~\ref{eq:update_theta} and $V^{(t)}$ using
        \Statex \quad~ Eq.~\ref{eq:update_V}, then send back to all devices.
		\EndFor
	\end{algorithmic}
\end{algorithm}
\vspace{-10pt}
\subsubsection{Outer Problem:} 
\label{sec:outer}
As shown in Fig.~\ref{fig:bilevel}, the outer problem~\ref{eq:outer} aims to refine the Koopman eigenfunctions $\Phi_{W_i}$ and reconstruction matrix $V_i$, utilizing the set of eigenvalues $\Lambda_i$ obtained from the optimized Koopman Operator $K_i^*$ in the inner problem. This focuses on minimizing the reconstruction errors between actual data points $y_i^{(j)} \in Y_i$ and their prediction derived through KMD in Eq.~\ref{eq:kmd} applied to an initial state $y_i^{(0)} \in Y_i$. Thus, helping the \textsc{ReKo} model adapt and respond to the evolving patterns over longer time frames in MVTS. 
To solve the outer problem, with the globally aggregated $K_i^{(t)}$, each device proceeds to optimize for $W_i{(t)}$, $V_i{(t)}$ using gradient-based methods (lines 9-11). The server then performs a second stage of aggregation, using Eq.~\ref{eq:update_theta} and ~\ref{eq:update_V} (line 12). 
\begin{align}
   W{(t)} & = \beta W^{(t-1)} + (1-\beta)\frac{1}{|\mathcal{S}^{(t)}|} \sum\nolimits_{i \in \mathcal{S}^{(t)}} W_i^{(t)} ,\label{eq:update_theta} \\ 
    V^{(t)} & = \beta V^{(t-1)} + (1-\beta)\frac{1}{|\mathcal{S}^{(t)}|} \sum\nolimits_{i \in \mathcal{S}^{(t)}} V_i^{(t)}, \label{eq:update_V}
\end{align}
where the role of \( \beta \) is similar to that of Eq.~\ref{eq:update_K}. 
This iterative cycle alternates between local optimizations and global aggregations for $T$ rounds, converging to a global \textsc{ReKo} model that captures the collective intelligence and unique attributes of all devices.

%% file: SDMSections_Full/algorithms.tex
\subsection{Computational Analysis:}
For \textit{time complexity}, the inner level involves gradient and norm calculations and spectral regularization with complexities $O(m^2b)$, $O(mb)$ (b is batch size), and $O(m^3)$, respectively, leading to $O(D(m^3 + 2m^2b))$ complexity for $D$ optimization iterations. The outer level also involves gradient and norm calculations and summing elements for $U$ optimization iterations, adding $O(U(L-s)(2mn^2 + n))$ complexity, where $L-s$ is the number of time steps. Totally, the overall complexity of \textsc{FedKO} over $T$ round is $O(T(D(m^3 + 2m^2b) + U(L-s)(2mn^2 + n)))$ for each client. 
For \textit{space complexity}, storing $\Phi$, $K$ and $V$ requires complexities of $O(mb)$ (b is batch size), $O(m^2)$ and $O(nm)$, respectively, totaling $O(m^2 + 2mn + mb)$.



%% file: SDMSections_Full/experiment.tex
\section{Experimental Results}
\label{sec:experiment}

\subsection{Experimental settings}

\subsubsection*{Datasets:}
We utilize four large-scale MVTS datasets: (1) \textit{Pool Server Metrics (PSM)}\cite{ransyncoder}, a 25-dimensional dataset from eBay's servers, sequentially distributed across 24 nodes using a Dirichlet distribution; (2) \textit{Server Machine Dataset (SMD)}\cite{omnianomaly}, containing five weeks of resource utilization data from 28 machines, each with 38 metrics like memory and CPU usage; (3) \textit{Soil Moisture Active Passive (SMAP)}\cite{smap}, provided by NASA's Mars rover, includes 55 entities each monitored by 25 sensors, capturing telemetry and soil samples. (4) \textit{Mars Science Laboratory (MSL)}\cite{smap}, also from NASA's Mars rover, collects data from 27 entities, each with 55 sensors. More details about these datasets and their non-IID characteristics are reported in Appendix~\ref{ap:datasets}.

\vspace{5pt}
\noindent\textbf{Baselines:} We compare \textsc{FedKO} against five SOTA baselines in MTAD: (1) \textit{DeepSVDD}~\cite{deepsvdd}, a deep neural network with Support Vector Data Description~\cite{SVDD}; (2) \textit{LSTM-AE}~\cite{lstmae}, a model combining LSTM units and Autoencoders to capture temporal dependencies; (3) \textit{USAD}~\cite{usad}, an adversarial encoder-decoder architecture designed for unsupervised MTAD; (4) \textit{GDN}~\cite{gdn}, a graph-based neural network for enhancing MTAD pattern recognition; (5) \textit{TranAD}~\cite{tranad}, a deep transformer networks for robust feature extraction in MVTS.
As these methods are typically applied in centralized settings, we employ FedAvg~\cite{fedavg} to enhance fairness and consistency when training them in FL environments.  Additionally, we assess the \textsc{ReKo} model on individual nodes with their local datasets (refer as \textit{Standalone}).

\vspace{5pt}
\noindent\textbf{Training Details:}  We standardize datasets using a z-score function before training. All models undergo 30 global rounds, each involving 25\% of devices randomly selected to train for 5 local epochs using Adam optimizers.For evaluation, we employ key metrics for identifying anomalies in MVTS including: (1) \textit{Precision}, (2) \textit{Recall}, (3) \textit{F1-Score}, and (4) the \textit{Area Under the Receiver Operating Characteristic curve} (AUC). In addition, we use the point-adjustment strategy~\cite{usad,tranad}, for calibrating Precision (Pre), Recall (Re), and F1-Score (F1) metrics. We provide detailed information about the datasets, baselines, training procedures, and reproducibility in Appendix~\ref{ap:experiment}.

\begin{table*}[t]
    \centering
    \caption{ The average evaluation metrics (\%) of \textsc{FedKO} and other models (with FedAvg) across MVTS datasets. Pre, Re and F1-Score are adjusted following~\cite{usad, tranad}. Metrics with best performance are \underline{underline}.}
    \label{tab:performance}
    \renewcommand{\arraystretch}{1.01} 
    \scalebox{0.84}{
        \begin{tabular}{|c|c||c|c||c|c|c|c|c|c|}
            
            \hline
            \multirow{2}{*}{\textbf{Dataset}} & \multirow{2}{*}{\textbf{Metric}} & \textbf{\textsc{ReKo}} & \multirow{2}{*}{\textbf{\textsc{FedKO}}} & \multicolumn{5}{c|}{\textbf{FedAvg}} 
            \\
            \cline{5-9}
            & & \textbf{Standalone} & 
            & \textbf{Deep SVDD} & \textbf{LSTM-AE} & \textbf{USAD} & \textbf{GDN}& \textbf{TranAD}
            \\

            \hline
            \hline
            \multirow{5}{*}{\textbf{PSM}} 
            & AUC       
            & 69.91 \footnotesize{$\pm$ 3.13}
            & \textit{\textbf{76.53}} \footnotesize{$\pm$ 0.60}  
            & 66.38 \footnotesize{$\pm$ 1.88} & 66.89 \footnotesize{$\pm$ 0.02} & 66.65 \footnotesize{$\pm$ 0.11} & \underline{{76.62}} \footnotesize{$\pm$ 1.21} & 62.74 \footnotesize{$\pm$ 0.28} \\
            & Pre       
            & 94.28 \footnotesize{$\pm$ 2.11} 
            & \underline{{\textbf{97.55}}} \footnotesize{$\pm$ 0.87}
            & 81.81 \footnotesize{$\pm$ 5.30} & 91.65 \footnotesize{$\pm$ 0.02} & 90.56 \footnotesize{$\pm$ 0.53} & 96.22 \footnotesize{$\pm$ 0.46} & 94.22 \footnotesize{$\pm$ 0.27} \\ 
            & Re    
            & 85.26 \footnotesize{$\pm$ 1.79} 
            & \underline{{\textbf{89.89}}} \footnotesize{$\pm$ 0.04}
            & 87.73 \footnotesize{$\pm$ 0.11} & 75.58 \footnotesize{$\pm$ 0.01} & 87.70 \footnotesize{$\pm$ 0.13} & 88.91 \footnotesize{$\pm$ 1.11} & 69.14 \footnotesize{$\pm$ 0.01} \\
            & F1       
            & 89.54 \footnotesize{$\pm$ 1.47} 
            & \underline{{\textbf{93.56}}} \footnotesize{$\pm$  0.41}
            & 84.63 \footnotesize{$\pm$ 2.89} & 82.84 \footnotesize{$\pm$ 0.02} & 89.07 \footnotesize{$\pm$ 0.20} & 92.53 \footnotesize{$\pm$ 0.11} & 79.81 \footnotesize{$\pm$ 0.09} \\ 
            \hline
            \multirow{5}{*}{\textbf{SMD}}
            & AUC      
            & 62.72 \footnotesize{$\pm$ 0.35} 
            & {\textbf{68.60}} \footnotesize{$\pm$ 0.08}
            & 58.92 \footnotesize{$\pm$ 0.12} & \underline{{68.89}} \footnotesize{$\pm$ 0.02} & 65.18 \footnotesize{$\pm$ 3.16} & 65.48 \footnotesize{$\pm$ 0.39} & 62.82 \footnotesize{$\pm$ 0.16} \\
            & Pre     
            & 75.70 \footnotesize{$\pm$ 0.21} 
            & \underline{\textit{\textbf{76.25}}} \footnotesize{$\pm$ 0.02}
            & 54.01 \footnotesize{$\pm$ 0.02} & 56.80 \footnotesize{$\pm$ 0.11} & 53.84 \footnotesize{$\pm$ 0.01} & 72.51 \footnotesize{$\pm$ 1.12} & 48.18 \footnotesize{$\pm$ 0.01} \\
            & Re  
            & 74.68 \footnotesize{$\pm$ 0.95} 
            & \underline{{\textbf{79.75}}} \footnotesize{$\pm$ 0.00} 
            & 57.33 \footnotesize{$\pm$ 4.70} & 54.17 \footnotesize{$\pm$ 0.36} & 70.42 \footnotesize{$\pm$ 0.01} & 62.85 \footnotesize{$\pm$ 0.97} & 49.50 \footnotesize{$\pm$ 0.01}\\
            & F1     
            & 74.56 \footnotesize{$\pm$ 0.02} 
            & \underline{{\textbf{77.95}}} \footnotesize{$\pm$ 0.01}
            & 62.62 \footnotesize{$\pm$ 4.90} & 55.49 \footnotesize{$\pm$ 0.24} & 61.02 \footnotesize{$\pm$ 0.01} & 67.37 \footnotesize{$\pm$ 1.29}  & 48.83 \footnotesize{$\pm$ 0.01} \\
            \hline
            \multirow{5}{*}{\textbf{SMAP}}
            & AUC     
            & 43.17 \footnotesize{$\pm$ 3.31}
            & \textbf{48.40} \footnotesize{$\pm$  0.50}
            & 47.75 \footnotesize{$\pm$ 8.90} & 37.91 \footnotesize{$\pm$ 0.67} &  40.23 \footnotesize{$\pm$ 0.91} & 51.92 \footnotesize{$\pm$ 1.33} & \underline{55.64} \footnotesize{$\pm$ 1.07} \\
            & Pre   
            & 91.45 \footnotesize{$\pm$ 0.52}
            & \textbf{94.67} \footnotesize{$\pm$ 0.10}
            & 89.86 \footnotesize{$\pm$ 1.54} & 84.72 \footnotesize{$\pm$ 1.47} & 92.36 \footnotesize{$\pm$ 0.01} & \underline{95.01} \footnotesize{$\pm$ 2.25} & 91.66 \footnotesize{$\pm$ 2.14}  \\
            & Re 
            & 57.10 \footnotesize{$\pm$ 0.24}
            & \underline{{\textbf{57.25}}} \footnotesize{$\pm$ 0.10}
            & 57.15 \footnotesize{$\pm$ 1.46} & 54.92 \footnotesize{$\pm$ 0.01} & 54.92 \footnotesize{$\pm$ 0.01} & 56.29 \footnotesize{$\pm$ 1.37} & 55.04 \footnotesize{$\pm$ 2.60} \\
            & F1  
            & 70.30 \footnotesize{$\pm$ 0.71}
            & \underline{{\textbf{71.35}}} \footnotesize{$\pm$ 0.10}
            & 71.30 \footnotesize{$\pm$ 0.14} & 66.64 \footnotesize{$\pm$ 0.50} & 68.88 \footnotesize{$\pm$ 0.01} & 70.57 \footnotesize{$\pm$ 0.38} & 68.78 \footnotesize{$\pm$ 2.04} \\
            \hline
            \multirow{5}{*}{\textbf{MSL}}
            & AUC  
            & 55.30 \footnotesize{$\pm$ 2.31}
            & \underline{{\textbf{56.74}}} \footnotesize{$\pm$ 1.00}
            & 56.33 \footnotesize{$\pm$ 0.23} & 54.37 \footnotesize{$\pm$ 1.13} & 55.65 \footnotesize{$\pm$ 0.70} & 55.35 \footnotesize{$\pm$ 0.69} & 55.24 \footnotesize{$\pm$ 1.67} \\
            & Pre   
            & 76.61 \footnotesize{$\pm$ 0.20}
            & \underline{{\textbf{83.61}}} \footnotesize{$\pm$ 0.03}            
            & 74.27 \footnotesize{$\pm$ 3.80} & 51.64 \footnotesize{$\pm$ 3.12} & 64.13 \footnotesize{$\pm$ 0.01} & 74.18 \footnotesize{$\pm$ 0.48} & 63.28 \footnotesize{$\pm$ 5.21} \\ 
            & Re  
            & 83.81 \footnotesize{$\pm$ 0.35} 
            & {\textbf{87.27}} \footnotesize{$\pm$ 0.00}
            & 86.86 \footnotesize{$\pm$ 0.80} & 72.05 \footnotesize{$\pm$ 0.01} & 71.90 \footnotesize{$\pm$ 0.01} & 86.92 \footnotesize{$\pm$ 1.50} &  \underline{92.46} \footnotesize{$\pm$ 4.63} \\
            & F1    
            & 79.40 \footnotesize{$\pm$ 0.21} 
            & \underline{{\textbf{85.40}}} \footnotesize{$\pm$ 0.01}
            & 80.03 \footnotesize{$\pm$ 2.55} & 60.13 \footnotesize{$\pm$ 2.12} & 67.80 \footnotesize{$\pm$ 0.01} & 80.05 \footnotesize{$\pm$ 0.92} & 74.83 \footnotesize{$\pm$ 2.20} \\
            \hline
        \end{tabular}
    }
\end{table*}

\subsection{Main Results}
\subsubsection*{Performance of \textsc{FedKO} on MTAD Tasks:}

\begin{figure}[t]
 	\vspace{-5pt}
	\centering    
	\includegraphics[width=0.8\linewidth]{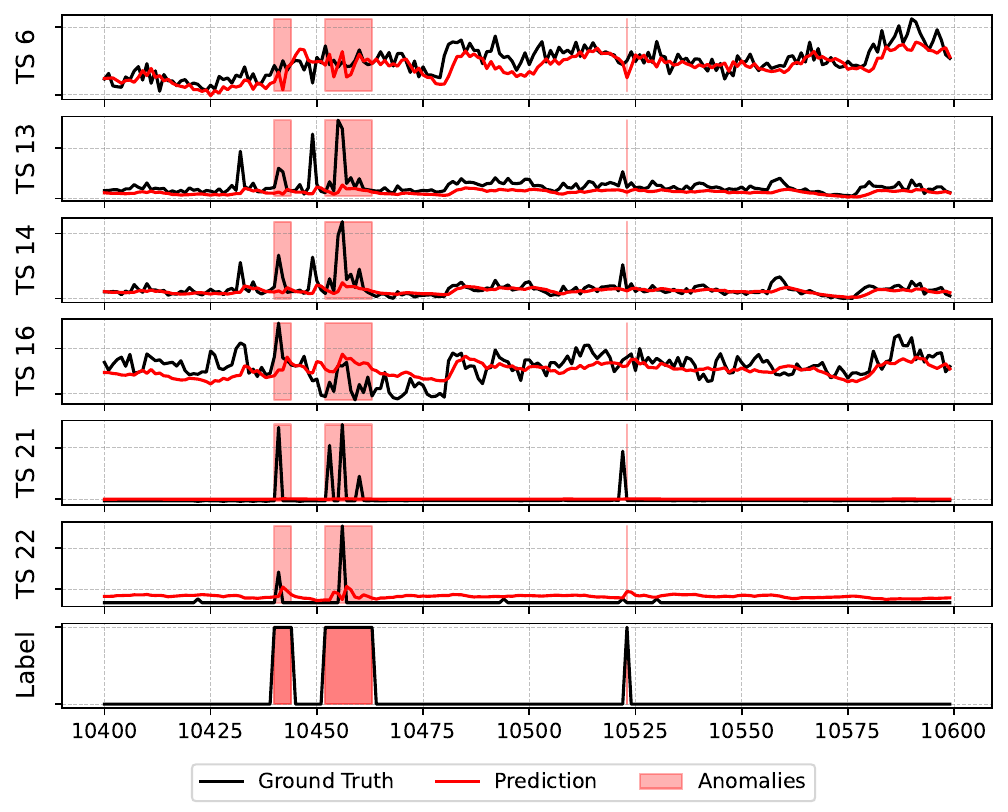}
	\caption{Prediction using \textsc{FedKO} on the PSM dataset}
	\label{fig:prediction_psm}
     \vspace{-10pt}
\end{figure}

We evaluate \textsc{FedKO} and benchmark its performance against established baselines in MTAD tasks. The results in Table~\ref{tab:performance} underscore \textsc{FedKO}'s superior capability in detecting anomalies with a significant margin of improvement over conventional models, demonstrating its ability to \textit{balance precision and recall} effectively. This high and balanced performance is crucial in operational settings, where accurately identifying normal and anomalous behaviors directly influences the reliability and efficiency of monitoring systems. 

In PSM, \textsc{FedKO} consistently surpasses its competitors across all metrics. Meanwhile, in SMD, it achieves a balanced precision and recall, leading to an F1-score that notably exceeds its counterparts. In SMAP, it also offers a better Recall and F1-score over advanced models like GDN and TranAD. This trend extends to the MSL dataset, where \textsc{FedKO} achieves the top performance. It is worth noting that sophisticated models like GDN and TranAD may perform well in centralized settings but could see reduced effectiveness in FL due to limited resources and data heterogeneity. Furthermore, DeepSVDD, LSTM\_AE, and USAD often struggle in FL and MVTS due to a lack of robust spatiotemporal modeling. The \textsc{ReKo} model shows effective performance even on individual datasets or other FL approaches, underscoring its value and responsiveness in data-scarce scenarios. 
 In Fig.~\ref{fig:prediction_psm}, we present \textsc{FedKO}'s predictions over time in the PSM dataset, showing its precision in mirroring the ground truth across various time series (TS). 
The alignment of \textsc{FedKO}'s red prediction line with the black ground truth line underscores its capability to accurately reflect data dynamics. The minimal shaded pink regions indicate \textsc{FedKO}'s effective anomaly detection without overfitting to noise, with these anomalies aligning closely to significant deviations in the ground truth. This showcases \textsc{FedKO}'s precise and reliable identification of unusual data patterns.

\begin{figure}[t]
	\centering
 	\begin{subfigure}{.51\linewidth}
		\centering
		\includegraphics[width=\textwidth]{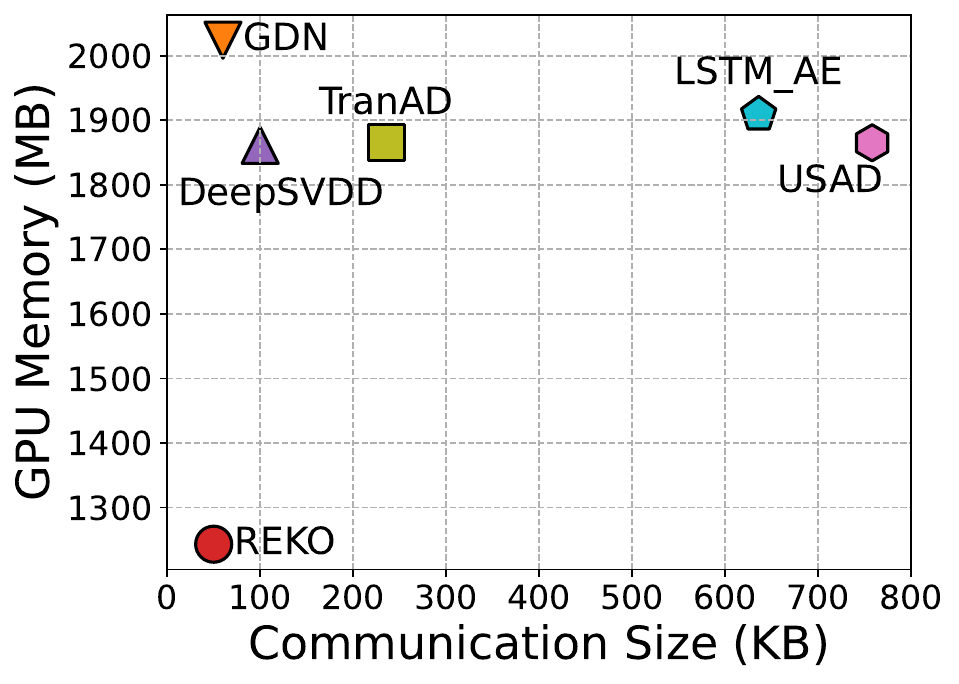}
        \vspace{-15pt}
		\caption{}
		\label{fig:model_size}
	\end{subfigure}%
	\begin{subfigure}{.49\linewidth}
		\centering
		\includegraphics[width=\textwidth]{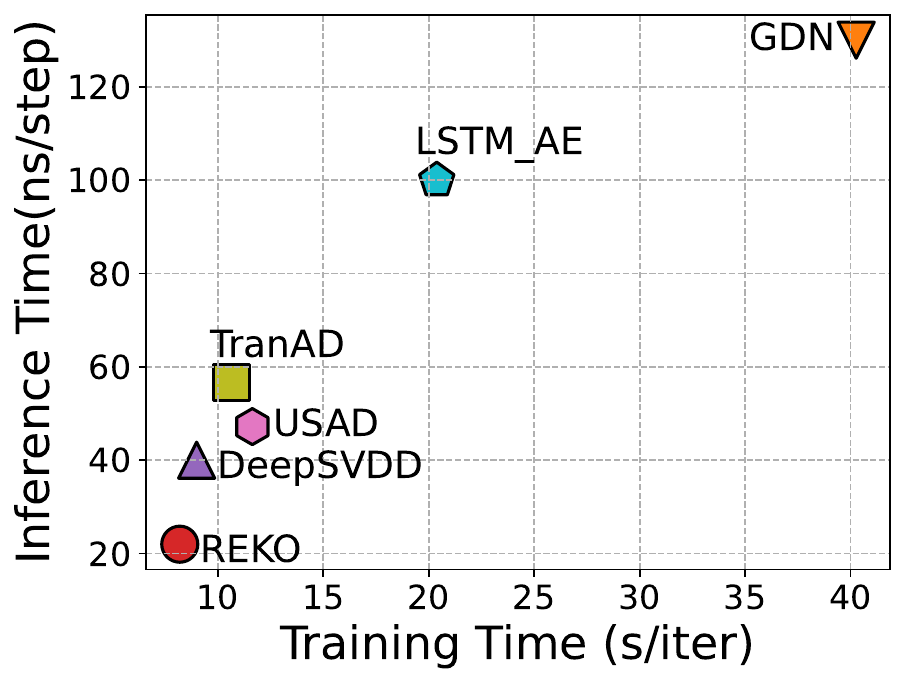}
        \vspace{-15pt}
		\caption{}
		\label{fig:time}
	\end{subfigure}
	\caption{(a) Communication Size vs Peak Memory Usage; (b) Training Time vs Inference Time}
    \label{fig:datasets}
	\vspace{-15pt}
\end{figure}
\vspace{5pt}
\noindent\textbf{Communication and Memory Efficiency:} We assessed the \textit{model size} transmission and the \textit{memory footprint} of each node within a single training round on the PSM dataset, aiming to benchmark \textsc{FedKO} against baseline models. As depicted in Fig.~\ref{fig:model_size}, \textsc{FedKO} stands out for its notably smaller model size and reduced memory requirements. In contrast, the USAD and LSTM\_AE models demand over 8 times larger sizes due to their deep architectures. While GDN models align closely with \textsc{FedKO}'s size, their memory needs exceed \textsc{FedKo} by approximately 2 times. Both DeepSVDD and Tran\_AD, despite having moderate sizes, still necessitate more memory demands than \textsc{FedKO} and show inferior performance. These underscore \textsc{FedKO}'s efficiency, showcasing its streamlined architecture that significantly improves communication and memory usage in resource-limited environments. 

\vspace{5pt}
\noindent\textbf{Training and Inference Time:}
We evaluated the time complexity of \textsc{ReKo} against baseline models, measuring training time per communication round and inference time per time step on a single node. As shown in Fig.~\ref{fig:time}, \textsc{FedKO} stands out for its efficiency in both training and inference, attributable to its inherent design that leverages RC to process MVTS. In comparison, models such as DeepSVDD and USAD, which employ simpler architectures, offered a middle ground in terms of both training and inference. Conversely, techniques like LSTM-AE, GDN, and TranAD, dependent on more intricate structure, face considerably longer training durations and increased latency in making predictions.

%% file: SDMSections_Full/conclusion.tex
\section{Conclusion}
\label{sec:conclusion}
We present \textsc{FedKO}, a resource-efficient FL framework for MVTS anomaly detection in large-scale systems. By combining the linear predictive power of KOT with the dynamic adaptability of RC within a bilevel optimization setup, \textsc{FedKO} effectively tackle the issues of data heterogeneity, variability, and privacy. Experimentally, \textsc{FedKO} demonstrates not only surpass performance over traditional MTAD methods, but also enhances communication and computational efficiency, appealing for large-scale applications.



%% file: SDMSections_Full/appendix1.tex
\newpage
\section{REPRODUCIBILITY}
\label{ap:experiment}
\subsection{Hardware and Software Packages}
In this study, all experiments are conducted on an AMD Ryzen 3970X Processor with 64 cores, 256GB of RAM, and four NVIDIA GeForce RTX 3090 GPUs. We use an coding environment with the main packages to benchmark and prototype as follows.
\begin{itemize}
    \item Python 3.10.12
    \item Pytorch 2.1.0+cu118
    \item Numpy
    \item Scipy
    \item Scikit-learn
\end{itemize}

\textbf{Memory Footprint:} To estimate GPU memory usage, we use the system-level \textit{nvidia-smi} command and the \textit{py3nvml} library, which is a Python 3 wrapper around the NVIDIA Management Library. The memory usage is monitoring during the training phases.

\begin{table}[h]
	\centering
	\caption{Dataset Statistics (NS and NN are the number of time-series and the number of FL nodes, respectively)}
	\label{tab:dataset}
    \scalebox{0.85}{
    \begin{tabular}{p{0.8cm}rrcp{0.3cm}p{0.3cm}rr}
    \hline 
    & \textbf{Train} & \textbf{Test} & \textbf{Anomalies}  & \textbf{NS} & \textbf{NN} & \textbf{Mean} & \textbf{Std}\\
    \hline 
    \textbf{SMAP} & 135183 & 427617 & 12.85 \%  & 25 & 55 & 2560 & 645 \\
    \textbf{MSL} & 58317 & 73729  & 10.53 \% & 55 & 27 &  2159 & 990\\
    \textbf{SMD} & 708405 & 708420 & 4.16 \% & 38 & 28 & 25300 & 2332 \\
    \textbf{PSM} & 132481 & 87841 & 27.75 \% & 25 & 24 & -- & -- \\
    \hline
    \end{tabular}}
\end{table}

\subsection{Datasets}
\label{ap:datasets}
In this study, we use four publicly available datasets. Their descriptions, sources, and implementation details are summarized below:

\begin{itemize}[leftmargin=10pt]
    \item \textbf{Server Machine Dataset (SMD)}~\cite{ransyncoder}: offers an extensive look into the health and performance of server machines, featuring an enormous dataset of 708,405 training samples and an equally sized testing set of 708,420 samples. Anomalies represent a smaller fraction of the data at 4.16\%, reflecting the real-world rarity of significant issues in server operations. The dataset encompasses 38 time-series and 28 feature nodes, including metrics like CPU usage, memory load, and network traffic. This breadth and depth make SMD a critical benchmark for evaluating MTAD algorithms in maintaining server reliability and performance.
    \item \textbf{Pool Server Metrics (PSM)}~\cite{omnianomaly}: involves metrics from pooled server resources or data centers, with a focus on monitoring and anomaly detection within these infrastructures. It includes 132,481 training samples and 87,841 testing samples, with an anomaly rate of 27.75\%. With 25 time-series, PSM challenges models to discern subtle anomalies from normal fluctuations in server metrics, providing a valuable testbed for advancing MTAD techniques in the domain of IT operations.
    \item \textbf{Soil Moisture Active Passive (SMAP) Dataset}~\cite{smap}: derived from NASA's Soil Moisture Active Passive satellite mission, is designed for anomaly detection in multivariate time-series data. It comprises a substantial collection of 135,183 training samples and a larger testing set of 427,617 samples. Anomalies constitute 12.85\% of the dataset, highlighting its utility in identifying unusual patterns amidst predominantly normal operational data. With 25 distinct time-series and 55 feature nodes, the dataset offers a rich variety of telemetry metrics, challenging models to accurately detect deviations that could signify important anomalies in soil moisture measurements and satellite operations.
    \item \textbf{Mars Science Laboratory (MSL) Dataset}~\cite{smap}:  originates from the Mars Science Laboratory mission, focusing on the Curiosity rover's telemetry data. It includes 58,317 training samples and 73,729 testing samples, with anomalies making up 10.53\% of the data. This dataset provides a unique challenge with its 55 time-series and 27 feature nodes, representing various operational parameters of the rover. The presence of anomalies in this dataset is crucial for testing anomaly detection models that could potentially identify malfunctions or significant events affecting the rover's mission on Mars.
\end{itemize}

\textbf{Non-IID Settings:} 
For the PSM dataset, we adopt a non-i.i.d partitioning approach by assigning time points to 24 clients following a Dirichlet distribution, a method that is widely recognized in simulating data heterogeneity for FL. 

Regarding SMD, it comprises data collected from 28 distinct server machines. In our experimental setting, each server machine's dataset is treated as an individual FL client. This approach directly mirrors the heterogeneity inherent in real-world deployments, where each server can exhibit unique operational characteristics.

Similarly, the SMAP and MSL datasets are derived from measurements taken by 55 and 27 entities, respectively. These sensors vary in their resolutions, further contributing to the data's heterogeneity. In our FL setup, each node is allocated a dataset collected by one of these entities, leveraging the nature heterogeneity in their resolutions.

General statistics of these datasets are reported in Table~\ref{tab:dataset}.

\subsubsection{Evaluation Metrics.} We employ key metrics for identifying anomalies in MVTS including: (1) \textit{Precision} (the accuracy of anomaly detection), (2) \textit{Recall} (the model's ability to identify all actual anomalies), (3) \textit{F1-Score} (a balance between Precision and Recall), and (4) the \textit{Area Under the Receiver Operating Characteristic curve} (AUC) to measures the overall performance of MTAD models. In addition, we incorporate a point-adjustment strategy, widely used in established frameworks~\cite{usad,tranad}, for calibrating Precision (Pre), Recall (Re), and F1-Score (F1) metrics.

\subsection{Hyperparameters and settings}

\begin{table*}[t]
\centering
\caption{Hyperparameters for Each Dataset}
\scalebox{0.9}{
\begin{tabular}{|l|c|c|c|c|}
\hline
\textbf{Parameter} & \textbf{PSM} & \textbf{SMD} & \textbf{SMAP} & \textbf{MSL} \\
\hline
\hline
Number of feature & 25 & 38 & 27 & 55 \\
Dimension of $K$ & 128 & 128 & 128 & 256 \\
\hline
Local epochs & 5 & 5 & 5 & 5 \\
Global rounds & 30 & 30 & 30 & 30 \\
\hline
Batch Size (Inner) & 512 & 512 & 512 & 512 \\
Batch Size (Outer) & 128 & 128 & 128 & 128 \\
Train Rate & 85\% & 85\% & 85\% & 85\% \\
Validate Rate & 15\% & 15\% & 15\% & 15\% \\
\hline
Reservoir Size & 256 & 256 & 256 & 256 \\
Leaky Rate $\alpha$ & 0.75 & 0.75 & 0.75 & 0.75 \\
Spectral Radius & 0.99 & 0.99 & 0.99 & 0.99 \\
Reservoir Initializer & Uniform & Uniform & Uniform & Uniform \\
\hline
Aggregation factor $\beta$ & 0.5 & 0.6 & 0.7 & 0.7 \\
\hline
Optimizer & Adam & Adam & Adam & Adam \\
Learning Rate & $1 \times 10^{-3}$ & $1 \times 10^{-3}$ & $1 \times 10^{-3}$ & $1 \times 10^{-3}$ \\
\hline
Regularization & $1 \times 10^{-4}$ & $1 \times 10^{-4}$ & $1 \times 10^{-4}$ & $1 \times 10^{-4}$ \\
\hline
\end{tabular}}
\label{tab:hyperparameters_datasets}
\end{table*}

We provide the configuration of hyperparameters and settings tailored for each dataset within our study in Table~\ref{tab:hyperparameters_datasets}.

\subsubsection{Training Details}  For all methods, we train a global model for local datasets, without inter-node data sharing. Before training, all features undergo standardization via a z-score function. Common settings include 30 global rounds, where each round includes a randomly selected 25\% of the total devices trained for 5 local epochs with Adam optimizers. The inner and outer batch sizes of 512 and 128 respectively, and a training-validation split of 85\%-15\%. The dimension of \(K\) is 128 for PSM, SMD, and SMAP, and 256 for MSL. For $\Phi$, we use Leaky ESN with a size of 256, a leaky rate of 0.75, a spectral radius of 0.99, and a uniform initializer. The aggregation factor \( \beta \) ranges from 0.5 to 0.7. For implementation, we use a coding environment with Python 3.10, Pytorch 2.1.0, and CUDA 12.1. Experiments are mainly conducted on an Intel® Xeon® W-3335 Server with 512GB RAM and NVIDIA RTX 4090 GPUs.
 
%

\subsection{\textsc{REKO} Model}
The implementation details for each component of \textsc{ReKo} model are detailed below:

\subsubsection{Reservoir-Koopman Lifted Linearization Component}
To implement $\Phi$, we employ Leaky Echo State Network (Leaky ESN)~\cite{esn}, a specialized form of Recurrent Neural Network (RNN) designed for efficient processing of sequential data through the paradigm of reservoir computing. 
Central to  Leaky ESN's design is the reservoir—a large, dynamically rich, yet fixed network of neurons generated randomly. This reservoir acts as a temporal kernel, transforming input sequences into a higher-dimension space where linear separation is feasible. It is mathematically represented as:
\[ r^{(t)} = (1-\alpha)r^{(t-1)} + \alpha f(W_{in}x^{(t)} + W_{res}r^{(t-1)} + b_{res}) \]
where:
\begin{itemize}
    \item \(r^{(t)}\) is the state of the reservoir neurons at time \(t\),
    \item \(\alpha\) is the leaking rate parameter (\(0< \alpha \leq 1\)), controlling the rate of leaking,
    \item \(f\) is a non-linear activation function,
    \item \(W_{in}\) are the input weights,
    \item \(x^{(t)}\) is the input at time \(t\),
    \item \(W_{res}\) are the recurrent weights within the reservoir,
    \item \(b_{res}\) is a bias term.
\end{itemize}

The term "leaky" denotes the integration of a leaky integrator in the neurons of the reservoir. Unlike the straightforward state update in standard ESNs, which involves a simple weighted sum of incoming signals activated by a non-linear function, Leaky ESNs introduce a leaky term to this equation. This term enables the neuron's state to gradually decay or "leak" towards a resting state before receiving new inputs. The leaky rate \(\alpha\) acts as a temporal scaling mechanism that enhances the network's adaptability and flexibility. By controlling this rate speed of the network's internal states, Leaky ESNs exhibit improved proficiency in capturing and modeling complex temporal dynamics. This characteristic renders them particularly effective for time-series prediction.

Within the architecture of a Leaky ESN, the readout layer, also known as the output layer, is responsible for transforming the complex, high-dimensional activity patterns of the reservoir into a format suitable for specific tasks, such as classification, regression, or prediction. This transformation is achieved through a learned linear or nonlinear mapping, depending on the task's requirements.
The functionality of the readout layer is predicated on the premise that the reservoir—enhanced by the leaky integrator mechanism—efficiently encodes temporal and spatial input features into its state dynamics. These dynamics are captured in the state vector \(r^{(t)}\), which aggregates the activations of the reservoir neurons at a given time \(t\).

The readout layer operates by applying a set of trainable weights, \(W_{out}\), to the reservoir's state vector. The output of the network at time \(t\), denoted by \(y(t)\), is calculated as follows:

\[ y^{(t)} = h(Wr^{(t)} + b) \]

where:
\begin{itemize}
    \item \(h\) is the activation function of the readout layer, which can be linear or nonlinear depending on the desired output characteristics of the network,
    \item \(W\) is the weight matrix connecting the reservoir states to the output units,
    \item \(b\) represents the bias term for the output layer.
\end{itemize}

The training of the readout layer in a Leaky ESN is conducted through a learning algorithm that adjusts \(W\) (and possibly \(b\), if included) to minimize the difference between the network's output and the target output. Commonly, this training process involves linear regression techniques such as least squares for tasks where a linear readout suffices, or more complex algorithms for tasks requiring nonlinear mappings.

A distinctive feature of the Leaky ESN is that the training process is confined to the readout layer. The reservoir remains fixed after initialization. This constraint significantly simplifies the training process, as it avoids the need for backpropagation through time (BPTT) or other computationally intensive techniques commonly used in traditional RNN training. The separation of the dynamic reservoir processing from the static, trainable readout layer allows for efficient training and adaptation of the network to a wide range of tasks, leveraging the rich temporal representations generated by the reservoir.

\subsubsection{Koopman Operator}

Within the \textsc{ReKo} model, the Koopman Operator, denoted as $K$, is conceptualized as a transformative matrix pivotal to modeling the dynamical system's evolution. To operationalize $K$ within our neural architecture, we implement it as a linear layer. This approach facilitates the learning of complex dynamics by enabling the model to approximate the Koopman Operator's action on the system's state space through linear transformations. By embodying $K$ in this manner, we leverage the inherent linearity of the Koopman Operator theory, allowing for a more efficient and interpretable representation of the system's temporal evolution.

\textbf{Spectral Radius Regularization:} The Koopman Operator offers a powerful linear perspective on nonlinear dynamical systems by operating on the infinite-dimensional space of observable functions. This operator's spectral properties, particularly the spectral radius, play a crucial role in understanding and predicting the system's dynamics. Regularizing the spectral radius of the Koopman Operator's approximations is essential for ensuring the stability and accuracy of the analysis. Given an approximation of the Koopman Operator $K$, its spectral radius $\rho(K)$ is determined by the largest absolute eigenvalue of $K$. To regularize $\rho(K)$ and ensure the stability of the dynamical system analysis, we apply the direct scaling method can be applied. This method adjusts $\rho(K)$ to a desired target value $\rho_{\text{target}}$, conducive to stable and meaningful dynamical analysis. The scaling is performed as follows:

\begin{equation}
    K_{\text{scaled}} = \frac{K}{\rho(K)} \times \rho_{\text{target}}
\end{equation}

where $K_{\text{scaled}}$ represents the Koopman Operator after scaling, ensuring that its spectral radius is adjusted to $\rho_{\text{target}}$, the preselected target spectral radius.

\subsubsection{Reconstruction Matrix}

Similarly, the reconstruction matrix $V$ in the \textsc{ReKo} model, is essential for mapping the transformed state space back to the original state dimensions. We instantiate $V$ as a linear layer within our model architecture, signifying its role in the learning process. This configuration enables the model to learn an optimal linear mapping that reconstructs the observed data from its Koopman-transformed representation. By defining $V$ through a linear layer, we ensure a direct and efficient mechanism for recovering the system's states, thereby facilitating a coherent integration between the Koopman Operator's abstract state transformations and the physical state reconstructions.

\subsubsection{Alternatives or enhancements to LESN:}
We choose the Leaky ESN in \textsc{FedKO} its with  for lifting linearization, facilitating linear prediction with the Koopman operator. There are several modern reservoir computing architectures that can be considered for this task, including next-generation reservoir computing (NG-RC)~\cite{ngrc} or Reservoir Transformers~\cite{rctran}, which we will study in future improvements.

\subsection{Code Availability}

The source code supporting our work will be published at the following link: 
\href{https://github.com/dual-grp/FedKO.git}{https://github.com/dual-grp/FedKO.git}



%% file: sdm_arxiv.bbl
\begin{thebibliography}{10}
\providecommand{\url}[1]{#1}
\csname url@samestyle\endcsname
\providecommand{\newblock}{\relax}
\providecommand{\bibinfo}[2]{#2}
\providecommand{\BIBentrySTDinterwordspacing}{\spaceskip=0pt\relax}
\providecommand{\BIBentryALTinterwordstretchfactor}{4}
\providecommand{\BIBentryALTinterwordspacing}{\spaceskip=\fontdimen2\font plus
\BIBentryALTinterwordstretchfactor\fontdimen3\font minus
  \fontdimen4\font\relax}
\providecommand{\BIBforeignlanguage}[2]{{%
\expandafter\ifx\csname l@#1\endcsname\relax
\typeout{** WARNING: IEEEtran.bst: No hyphenation pattern has been}%
\typeout{** loaded for the language `#1'. Using the pattern for}%
\typeout{** the default language instead.}%
\else
\language=\csname l@#1\endcsname
\fi
#2}}
\providecommand{\BIBdecl}{\relax}
\BIBdecl

\bibitem{arima}
\BIBentryALTinterwordspacing
G.~Box and G.~Jenkins, \emph{Time {Series} {Analysis}: {Forecasting} and
  {Control}}, ser. Holden-{Day} series in time series analysis and digital
  processing.\hskip 1em plus 0.5em minus 0.4em\relax Holden-Day, 1970.
  \url{https://books.google.com.au/books?id=5BVfnXaq03oC}
\BIBentrySTDinterwordspacing

\bibitem{smap}
\BIBentryALTinterwordspacing
K.~Hundman, V.~Constantinou, C.~Laporte, I.~Colwell, and T.~Soderstrom,
  ``Detecting spacecraft anomalies using lstms and nonparametric dynamic
  thresholding,'' in \emph{Proceedings of the 24th ACM SIGKDD International
  Conference on Knowledge Discovery \& Data Mining}, 2018, p. 387–395.
  \url{https://doi.org/10.1145/3219819.3219845}
\BIBentrySTDinterwordspacing

\bibitem{lstmae}
\BIBentryALTinterwordspacing
A.~Garg, W.~Zhang, J.~Samaran, R.~Savitha, and C.-S. Foo, ``An evaluation of
  anomaly detection and diagnosis in multivariate time series,'' \emph{IEEE
  Transactions on Neural Networks and Learning Systems}, vol.~33, pp.
  2508--2517, 2021.  \url{https://api.semanticscholar.org/CorpusID:80861784}
\BIBentrySTDinterwordspacing

\bibitem{vae1}
\BIBentryALTinterwordspacing
L.~Li, J.~Yan, H.~Wang, and Y.~Jin, ``Anomaly detection of time series with
  smoothness-inducing sequential variational auto-encoder,'' \emph{IEEE
  Transactions on Neural Networks and Learning Systems}, vol.~32, pp.
  1177--1191, 2020.  \url{https://api.semanticscholar.org/CorpusID:215773891}
\BIBentrySTDinterwordspacing

\bibitem{ransyncoder}
\BIBentryALTinterwordspacing
A.~Abdulaal, Z.~Liu, and T.~Lancewicki, ``Practical approach to asynchronous
  multivariate time series anomaly detection and localization,'' in
  \emph{Proceedings of the 27th ACM SIGKDD Conference on Knowledge Discovery \&
  Data Mining}, 2021, p. 2485–2494.
  \url{https://doi.org/10.1145/3447548.3467174}
\BIBentrySTDinterwordspacing

\bibitem{ae_sdm}
\BIBentryALTinterwordspacing
K.-H. Lai, L.~Wang, H.~Chen, K.~Zhou, F.~Wang, H.~Yang, and X.~Hu,
  \emph{Context-aware Domain Adaptation for Time Series Anomaly Detection}, pp.
  676--684.  \url{https://epubs.siam.org/doi/abs/10.1137/1.9781611977653.ch76}
\BIBentrySTDinterwordspacing

\bibitem{gan1}
\BIBentryALTinterwordspacing
D.~Li, D.~Chen, B.~Jin, L.~Shi, J.~Goh, and S.-K. Ng, ``Mad-gan: Multivariate
  anomaly detection for time series data with generative adversarial
  networks,'' in \emph{Artificial Neural Networks and Machine Learning –
  ICANN 2019: Text and Time Series: 28th International Conference on Artificial
  Neural Networks, Proceedings, Part IV}, 2019, p. 703–716.
  \url{https://doi.org/10.1007/978-3-030-30490-4_56}
\BIBentrySTDinterwordspacing

\bibitem{usad}
\BIBentryALTinterwordspacing
J.~Audibert, P.~Michiardi, F.~Guyard, S.~Marti, and M.~A. Zuluaga, ``Usad:
  Unsupervised anomaly detection on multivariate time series,'' in
  \emph{Proceedings of the 26th ACM SIGKDD International Conference on
  Knowledge Discovery \& Data Mining}, 2020, p. 3395–3404.
  \url{https://doi.org/10.1145/3394486.3403392}
\BIBentrySTDinterwordspacing

\bibitem{daemon}
X.~Chen, L.~Deng, F.~Huang, C.~Zhang, Z.~Zhang, Y.~Zhao, and K.~Zheng,
  ``Daemon: Unsupervised anomaly detection and interpretation for multivariate
  time series,'' in \emph{2021 IEEE 37th International Conference on Data
  Engineering}, 2021, pp. 2225--2230.

\bibitem{gdn}
\BIBentryALTinterwordspacing
A.~Deng and B.~Hooi, ``Graph {Neural} {Network}-{Based} {Anomaly} {Detection}
  in {Multivariate} {Time} {Series},'' \emph{Proceedings of the AAAI Conference
  on Artificial Intelligence}, vol.~35, no.~5, pp. 4027--4035, May 2021.
  \url{https://ojs.aaai.org/index.php/AAAI/article/view/16523}
\BIBentrySTDinterwordspacing

\bibitem{gnn_icdm}
\BIBentryALTinterwordspacing
H.~Zhao, Y.~Wang, J.~Duan, C.~Huang, D.~Cao, Y.~Tong, B.~Xu, J.~Bai, J.~Tong,
  and Q.~Zhang, ``Multivariate time-series anomaly detection via graph
  attention network,'' in \emph{2020 IEEE International Conference on Data
  Mining (ICDM)}.\hskip 1em plus 0.5em minus 0.4em\relax Los Alamitos, CA, USA:
  IEEE Computer Society, nov 2020, pp. 841--850.
  \url{https://doi.ieeecomputersociety.org/10.1109/ICDM50108.2020.00093}
\BIBentrySTDinterwordspacing

\bibitem{grelen}
\BIBentryALTinterwordspacing
W.~Zhang, C.~Zhang, and F.~Tsung, ``Grelen: Multivariate time series anomaly
  detection from the perspective of graph relational learning,'' in
  \emph{Proceedings of the Thirty-First International Joint Conference on
  Artificial Intelligence, {IJCAI-22}}, 7 2022, pp. 2390--2397.
  \url{https://doi.org/10.24963/ijcai.2022/332}
\BIBentrySTDinterwordspacing

\bibitem{tranad}
\BIBentryALTinterwordspacing
S.~Tuli, G.~Casale, and N.~R. Jennings, ``Tranad: Deep transformer networks for
  anomaly detection in multivariate time series data,'' \emph{Proc. VLDB
  Endow.}, vol.~15, no.~6, p. 1201–1214, feb 2022.
  \url{https://doi.org/10.14778/3514061.3514067}
\BIBentrySTDinterwordspacing

\bibitem{anomaly_transformer}
\BIBentryALTinterwordspacing
J.~Xu, H.~Wu, J.~Wang, and M.~Long, ``Anomaly transformer: Time series anomaly
  detection with association discrepancy,'' in \emph{International Conference
  on Learning Representations}, 2022.
  \url{https://openreview.net/forum?id=LzQQ89U1qm_}
\BIBentrySTDinterwordspacing

\bibitem{fedavg}
\BIBentryALTinterwordspacing
H.~B. McMahan, E.~Moore, D.~Ramage, S.~Hampson, and B.~A.~y. Arcas,
  ``Communication-efficient learning of deep networks from decentralized
  data,'' 2016.  \url{https://arxiv.org/abs/1602.05629}
\BIBentrySTDinterwordspacing

\bibitem{koopman}
\BIBentryALTinterwordspacing
B.~O. Koopman, ``Hamiltonian systems and transformations in hilbert space,''
  \emph{Proceedings of the National Academy of Sciences of the United States of
  America}, vol.~17, no.~5, pp. 315--318, 1931.
  \url{http://www.jstor.org/stable/86114}
\BIBentrySTDinterwordspacing

\bibitem{reservoir}
\BIBentryALTinterwordspacing
K.~Nakajima and I.~Fischer, Eds., \emph{\BIBforeignlanguage{en}{Reservoir
  {Computing}: {Theory}, {Physical} {Implementations}, and {Applications}}},
  ser. Natural {Computing} {Series}.\hskip 1em plus 0.5em minus 0.4em\relax
  Singapore: Springer, 2021.
  \url{https://link.springer.com/10.1007/978-981-13-1687-6}
\BIBentrySTDinterwordspacing

\bibitem{Konecny2015}
\BIBentryALTinterwordspacing
J.~Konečný, B.~McMahan, and D.~Ramage, ``Federated optimization:distributed
  optimization beyond the datacenter,'' 2015.
  \url{https://arxiv.org/abs/1511.03575}
\BIBentrySTDinterwordspacing

\bibitem{fl_iot}
D.~C. Nguyen, M.~Ding, P.~N. Pathirana, A.~Seneviratne, J.~Li, and H.~V. Poor,
  ``Federated learning for internet of things: A comprehensive survey,''
  \emph{{IEEE} Communications Surveys {\&} Tutorials}, vol.~23, no.~3, pp.
  1622--1658, 2021.

\bibitem{fl_iot_sdm}
\BIBentryALTinterwordspacing
J.~Wang, S.~Zeng, Z.~Long, Y.~Wang, H.~Xiao, and F.~Ma,
  \emph{Knowledge-Enhanced Semi-Supervised Federated Learning for Aggregating
  Heterogeneous Lightweight Clients in IoT}, pp. 496--504.
  \url{https://epubs.siam.org/doi/abs/10.1137/1.9781611977653.ch56}
\BIBentrySTDinterwordspacing

\bibitem{flames2graph}
\BIBentryALTinterwordspacing
R.~Younis, Z.~Ahmadi, A.~Hakmeh, and M.~Fisichella, ``Flames2graph: An
  interpretable federated multivariate time series classification framework,''
  in \emph{Proceedings of the 29th ACM SIGKDD Conference on Knowledge Discovery
  and Data Mining}, 2023, p. 3140–3150.
  \url{https://doi.org/10.1145/3580305.3599354}
\BIBentrySTDinterwordspacing

\bibitem{fedmssa}
\BIBentryALTinterwordspacing
J.~He, M.~Khushi, T.~Nguyen, and N.~H. Tran, ``Fed-mssa: A federated approach
  for spatio-temporal data modeling using multivariate singular spectrum
  analysis,'' in \emph{2023 IEEE International Conference on Data Mining
  (ICDM)}.\hskip 1em plus 0.5em minus 0.4em\relax IEEE Computer Society, dec
  2023, pp. 1055--1060.
  \url{https://doi.ieeecomputersociety.org/10.1109/ICDM58522.2023.00123}
\BIBentrySTDinterwordspacing

\bibitem{pmlr-v108-reisizadeh20a}
\BIBentryALTinterwordspacing
A.~Reisizadeh, A.~Mokhtari, H.~Hassani, A.~Jadbabaie, and R.~Pedarsani,
  ``Fedpaq: A communication-efficient federated learning method with periodic
  averaging and quantization,'' in \emph{Proceedings of the Twenty Third
  International Conference on Artificial Intelligence and Statistics}, vol.
  108, 2020, pp. 2021--2031.
  \url{https://proceedings.mlr.press/v108/reisizadeh20a.html}
\BIBentrySTDinterwordspacing

\bibitem{com1}
F.~Sattler, S.~Wiedemann, K.-R. Müller, and W.~Samek, ``Robust and
  communication-efficient federated learning from non-i.i.d. data,'' \emph{IEEE
  Transactions on Neural Networks and Learning Systems}, vol.~31, no.~9, pp.
  3400--3413, 2020.

\bibitem{comadaptive}
\BIBentryALTinterwordspacing
Y.~Wang, L.~Lin, and J.~Chen, ``Communication-efficient adaptive federated
  learning,'' in \emph{Proceedings of the 39th International Conference on
  Machine Learning}, ser. Proceedings of Machine Learning Research, vol.
  162.\hskip 1em plus 0.5em minus 0.4em\relax PMLR, 17--23 Jul 2022, pp.
  22\,802--22\,838.  \url{https://proceedings.mlr.press/v162/wang22o.html}
\BIBentrySTDinterwordspacing

\bibitem{fluid_dynamics}
\BIBentryALTinterwordspacing
I.~Mezi\'{c}, ``Analysis of fluid flows via spectral properties of the koopman
  operator,'' \emph{Annual Review of Fluid Mechanics}, vol.~45, no.~1, pp.
  357--378, 2013.  \url{https://doi.org/10.1146/annurev-fluid-011212-140652}
\BIBentrySTDinterwordspacing

\bibitem{kotop}
\BIBentryALTinterwordspacing
E.~Kaiser, J.~N. Kutz, and S.~L. Brunton, ``Data-driven discovery of koopman
  eigenfunctions for control,'' \emph{Machine Learning: Science and
  Technology}, vol.~2, 2017.
  \url{https://api.semanticscholar.org/CorpusID:49216843}
\BIBentrySTDinterwordspacing

\bibitem{dmd1}
\BIBentryALTinterwordspacing
P.~J. Schmid and P.~Ecole, ``Dynamic mode decomposition of numerical and
  experimental data,'' \emph{Journal of Fluid Mechanics}, vol. 656, pp. 5 --
  28, 2008.  \url{https://api.semanticscholar.org/CorpusID:11334986}
\BIBentrySTDinterwordspacing

\bibitem{dmd2}
\BIBentryALTinterwordspacing
J.~H. Tu, C.~W. Rowley, D.~M. Luchtenburg, S.~L. Brunton, and J.~N. Kutz, ``On
  dynamic mode decomposition: Theory and applications,'' \emph{ACM Journal of
  Computer Documentation}, vol.~1, pp. 391--421, 2013.
  \url{https://api.semanticscholar.org/CorpusID:260502478}
\BIBentrySTDinterwordspacing

\bibitem{edmd1}
\BIBentryALTinterwordspacing
M.~O. Williams, I.~G. Kevrekidis, and C.~W. Rowley, ``A data–driven
  approximation of the koopman operator: Extending dynamic mode
  decomposition,'' \emph{Journal of Nonlinear Science}, vol.~25, pp. 1307 --
  1346, 2014.  \url{https://api.semanticscholar.org/CorpusID:5750630}
\BIBentrySTDinterwordspacing

\bibitem{edmd2}
\BIBentryALTinterwordspacing
Q.~Li, F.~Dietrich, E.~M. Bollt, and I.~G. Kevrekidis, ``Extended dynamic mode
  decomposition with dictionary learning: A data-driven adaptive spectral
  decomposition of the koopman operator.'' \emph{Chaos}, vol. 27 10, p. 103111,
  2017.  \url{https://api.semanticscholar.org/CorpusID:41957686}
\BIBentrySTDinterwordspacing

\bibitem{aekoopman}
\BIBentryALTinterwordspacing
B.~Lusch, J.~N. Kutz, and S.~L. Brunton, ``Deep learning for universal linear
  embeddings of nonlinear dynamics,'' \emph{Nature Communications}, vol.~9,
  2017.  \url{https://api.semanticscholar.org/CorpusID:4854885}
\BIBentrySTDinterwordspacing

\bibitem{aekoopman2}
N.~Takeishi, Y.~Kawahara, and T.~Yairi, ``Learning koopman invariant subspaces
  for dynamic mode decomposition,'' in \emph{Proceedings of the 31st
  International Conference on Neural Information Processing Systems}, 2017, p.
  1130–1140.

\bibitem{kotop1}
\BIBentryALTinterwordspacing
M.~Korda and I.~Mezi{\'c}, ``Optimal construction of koopman eigenfunctions for
  prediction and control,'' \emph{IEEE Transactions on Automatic Control},
  vol.~65, pp. 5114--5129, 2018.
  \url{https://api.semanticscholar.org/CorpusID:201107095}
\BIBentrySTDinterwordspacing

\bibitem{kotop2}
\BIBentryALTinterwordspacing
P.~Bevanda, J.~Kirmayr, S.~Sosnowski, and S.~Hirche, ``Learning the koopman
  eigendecomposition: A diffeomorphic approach,'' \emph{2022 American Control
  Conference (ACC)}, pp. 2736--2741, 2021.
  \url{https://api.semanticscholar.org/CorpusID:239009622}
\BIBentrySTDinterwordspacing

\bibitem{esn}
\BIBentryALTinterwordspacing
H.~Jaeger, ``The''echo state''approach to analysing and training recurrent
  neural networks,'' 2001.
  \url{https://api.semanticscholar.org/CorpusID:15467150}
\BIBentrySTDinterwordspacing

\bibitem{reservoir2}
F.~M. Bianchi, S.~Scardapane, S.~Løkse, and R.~Jenssen, ``Reservoir computing
  approaches for representation and classification of multivariate time
  series,'' \emph{IEEE Transactions on Neural Networks and Learning Systems},
  vol.~32, no.~5, pp. 2169--2179, 2021.

\bibitem{rctran}
\BIBentryALTinterwordspacing
S.~Shen, A.~Baevski, A.~Morcos, K.~Keutzer, M.~Auli, and D.~Kiela, ``Reservoir
  transformers,'' in \emph{Proceedings of the 59th Annual Meeting of the
  Association for Computational Linguistics and the 11th International Joint
  Conference on Natural Language Processing (Volume 1: Long Papers)}.\hskip 1em
  plus 0.5em minus 0.4em\relax Online: Association for Computational
  Linguistics, Aug. 2021, pp. 4294--4309.
  \url{https://aclanthology.org/2021.acl-long.331}
\BIBentrySTDinterwordspacing

\bibitem{optimalRC}
\BIBentryALTinterwordspacing
A.~Griffith, A.~Pomerance, and D.~J. Gauthier, ``Forecasting chaotic systems
  with very low connectivity reservoir computers,'' \emph{Chaos: An
  Interdisciplinary Journal of Nonlinear Science}, vol.~29, no.~12, p. 123108,
  Dec. 2019.  \url{https://doi.org/10.1063/1.5120710}
\BIBentrySTDinterwordspacing

\bibitem{optimalRCE}
\BIBentryALTinterwordspacing
L.~F. Livi, F.~M. Bianchi, and C.~Alippi, ``Determination of the edge of
  criticality in echo state networks through fisher information maximization,''
  \emph{IEEE Transactions on Neural Networks and Learning Systems}, vol.~29,
  pp. 706--717, 2016.  \url{https://api.semanticscholar.org/CorpusID:3575418}
\BIBentrySTDinterwordspacing

\bibitem{optimalRCG}
\BIBentryALTinterwordspacing
L.~A. Thiede and U.~Parlitz, ``Gradient based hyperparameter optimization in
  {Echo} {State} {Networks},'' \emph{Neural Networks}, vol. 115, pp. 23--29,
  2019.
  \url{https://www.sciencedirect.com/science/article/pii/S0893608019300413}
\BIBentrySTDinterwordspacing

\bibitem{omnianomaly}
\BIBentryALTinterwordspacing
Y.~Su, Y.~Zhao, C.~Niu, R.~Liu, W.~Sun, and D.~Pei, ``Robust anomaly detection
  for multivariate time series through stochastic recurrent neural network,''
  in \emph{Proceedings of the 25th ACM SIGKDD International Conference on
  Knowledge Discovery \& Data Mining}, ser. KDD '19.\hskip 1em plus 0.5em minus
  0.4em\relax New York, NY, USA: Association for Computing Machinery, 2019, p.
  2828–2837.  \url{https://doi.org/10.1145/3292500.3330672}
\BIBentrySTDinterwordspacing

\bibitem{deepsvdd}
\BIBentryALTinterwordspacing
L.~Ruff, R.~Vandermeulen, N.~Goernitz, L.~Deecke, S.~A. Siddiqui, A.~Binder,
  E.~M{\"u}ller, and M.~Kloft, ``Deep one-class classification,'' in
  \emph{Proceedings of the 35th International Conference on Machine Learning},
  2018, pp. 4393--4402.  \url{https://proceedings.mlr.press/v80/ruff18a.html}
\BIBentrySTDinterwordspacing

\bibitem{SVDD}
D.~Tax and R.~Duin, ``Support vector data description,'' \emph{Machine
  Learning}, vol.~54, pp. 45--66, 01 2004.

\bibitem{ngrc}
\BIBentryALTinterwordspacing
D.~J. Gauthier, E.~Bollt, A.~Griffith, and W.~A.~S. Barbosa, ``Next generation
  reservoir computing,'' \emph{Nature Communications}, vol.~12, no.~1, p. 5564,
  Sep. 2021.  \url{https://doi.org/10.1038/s41467-021-25801-2}
\BIBentrySTDinterwordspacing

\end{thebibliography}
